\documentclass[10pt,twocolumn,letterpaper]{article}

\usepackage[pagenumbers]{cvpr} % To force page numbers, e.g. for an arXiv version

\usepackage{graphicx}
\usepackage{amsmath}
\usepackage{amssymb}
\usepackage{booktabs}

\usepackage[pagebackref,breaklinks,colorlinks]{hyperref}

\usepackage[dvipsnames,table,xcdraw]{xcolor}

\makeatletter
\@namedef{ver@everyshi.sty}{}
\makeatother
\usepackage{tikz}
\usepackage{color}
\usepackage{placeins}

\usepackage{array}
\usepackage{multirow}
\usepackage[flushleft]{threeparttable}
\usepackage{comment}
\usepackage{algorithmic}
\usepackage{algorithm}
\usepackage{booktabs}

\usepackage{placeins}
\usepackage{float}

\usepackage{amssymb}% http://ctan.org/pkg/amssymb
\usepackage{pifont}% http://ctan.org/pkg/pifont

\usepackage{fancyvrb}

\usepackage{graphicx}
\usepackage{enumitem}
\usepackage{wrapfig}
\usepackage{lipsum}

\usepackage{soul}
\usepackage[bb=dsserif]{mathalpha}
\usepackage{bm}

\usepackage{makecell}

\usepackage[capitalize]{cleveref}
\crefname{section}{Sec.}{Secs.}
\Crefname{section}{Section}{Sections}
\Crefname{table}{Table}{Tables}
\crefname{table}{Tab.}{Tabs.}

\newcommand{\cmark}{\text{\ding{51}}}%
\newcommand{\xmark}{\text{\ding{55}}}%

\newcommand{\ol}[3]{\begin{#1}[leftmargin=*,topsep=0pt]\setlength{\itemsep}{#2mm}#3\end{#1}}

\newlength\savewidth\newcommand\shline{\noalign{\global\savewidth\arrayrulewidth
  \global\arrayrulewidth 1pt}\hline\noalign{\global\arrayrulewidth\savewidth}}

\def\fontsmall#1#{
    \fontsize{8}{12}
    #1
    \selectfont
}

\def\tablefontsmall#1#{
    \fontsize{8}{10}
    #1
    \selectfont
}

\def\tablefont#1#{
    \fontsize{9}{12}
    #1
    \selectfont
}

\def\gray#1{\textcolor{gray}{#1}}

\usepackage[pagebackref,breaklinks,colorlinks]{hyperref}

\def\cvprPaperID{4116} % *** Enter the CVPR Paper ID here
\def\confName{CVPR}
\def\confYear{2023}

\begin{document}

%%%%%%%%% TITLE - PLEASE UPDATE
\title{Improving Unsupervised Video Object Segmentation\\
with Motion-Appearance Synergy}

% \author{First Author\\
% Institution1\\
% Institution1 address\\
% {\tt\small firstauthor@i1.org}
% % For a paper whose authors are all at the same institution,
% % omit the following lines up until the closing ``}''.
% % Additional authors and addresses can be added with ``\and'',
% % just like the second author.
% % To save space, use either the email address or home page, not both
% \and
% Second Author\\
% Institution2\\
% First line of institution2 address\\
% {\tt\small secondauthor@i2.org}
% }

% \author{
% \tb{ccc}{19}{
% Long Lian$^1$ &
% Zhirong Wu$^2$ &
% Stella X. Yu$^{1,3}$ \\
% }\\
% \tb{@{}ccc@{}}{10.5}{
% UC Berkeley/ICSI$^1$ & 
% Microsoft Research Asia$^2$ &
% University of Michigan$^3$
% }\\
% \tb{@{}ccc@{}}{10.5}{
% {\tt\small longlian@berkeley.edu} &
% {\tt\small wuzhiron@microsoft.com} &
% {\tt\small stellayu@umich.edu}
% }\\
% }

\author{
Long Lian$^\text{1}$, 
Zhirong Wu$^\text{2}$, 
Stella X. Yu$^\text{1,3}$ \\
$^\text{1}$UC Berkeley / ICSI, 
$^\text{2}$Microsoft Research Asia, 
$^\text{3}$University of Michigan \\
{\tt\small longlian@berkeley.edu, 
wuzhiron@microsoft.com, 
stellayu@umich.edu}
}
\maketitle

%%%%%%%%% ABSTRACT
\begin{abstract}
We present IMAS, a method that segments the primary objects in videos without manual annotation in training or inference. Previous methods in unsupervised video object segmentation (UVOS) have demonstrated the effectiveness of motion as either input or supervision for segmentation. However, motion signals may be uninformative or even misleading in cases such as deformable objects and objects with reflections, causing unsatisfactory segmentation.

In contrast, IMAS achieves \underline{I}mproved UVOS with \underline{M}otion-\underline{A}ppearance \underline{S}ynergy. Our method has two training stages: \textbf{1)}~a motion-supervised object discovery stage that deals with motion-appearance conflicts through a learnable residual pathway; \textbf{2)}~a refinement stage with both low- and high-level appearance supervision to correct model misconceptions learned from misleading motion cues.

Additionally, we propose motion-semantic alignment as a model-agnostic annotation-free hyperparam tuning method. We demonstrate its effectiveness in tuning critical hyperparams previously tuned with human annotation or hand-crafted hyperparam-specific metrics.

IMAS greatly improves the segmentation quality on several common UVOS benchmarks. For example, we surpass previous methods by 8.3\% on DAVIS16 benchmark with only standard ResNet and convolutional heads. We intend to release our code for future research and applications.
\end{abstract}

\def\tabAdvantage#1{
\begin{table}[#1]
\centering
%\vspace{-5pt}
\setlength{\tabcolsep}{1.8pt}
\begin{tabular}{l@{}cccc}
\shline
Method & \!\!MG {\small \cite{yang2021self}} & AMD {\small \cite{liu2021emergence}} & GWM {\small \cite{choudhury2022guess}} & \textbf{IMAS} \\
\shline
{\small Segment static objects} & \color{Red}{\xmark} & \color{Green}{\cmark} & \color{Green}{\cmark} & \color{Green}{\cmark} \\
{\small Supervision} & M & M$^*$ & M & M + A \\
{\small Learnable motion model} & -- & \color{Red}{\xmark} & \color{Red}{\xmark} & \color{Green}{\cmark} \\
\shline
\end{tabular}
\caption{\textbf{Key advantages of IMAS}: \textbf{1)} Using motion as supervision rather than input, IMAS is able to segment static objects. \textbf{2)} Using both motion (\textbf{M}) and appearance (\textbf{A}) as explicit supervision, IMAS is able to learn high-quality segmentation even from misleading motion signals. \textbf{3)} Learnable motion model allows implicit appearance learning in presence of motion-appearance cue conflicts. $*$: capturing segment-level motion by image warping.\vspace{-15pt}}
\label{tab:advantage}
\end{table}
}

\def\figTeaser#1{
\begin{figure}[#1]
\begin{subfigure}{0.39\linewidth}
\hspace{-10pt}
\includegraphics[width=\linewidth]{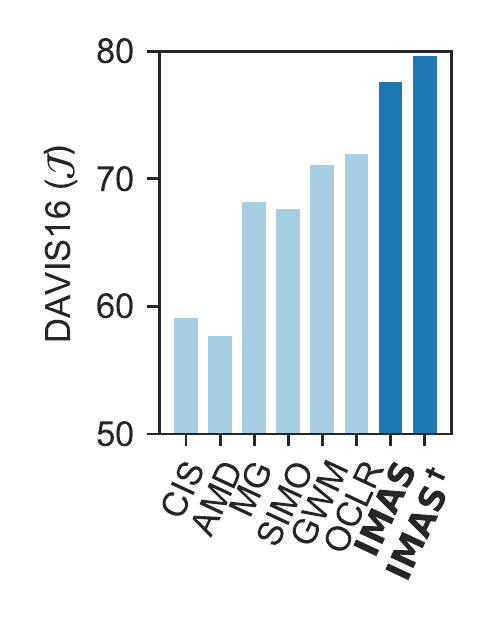}
\vspace{-12pt}
\caption{
Performance comparison
\vspace{-2pt}
}
\end{subfigure} \hfill
\begin{subfigure}{0.6\linewidth}
\centerline{\includegraphics[width=\linewidth]{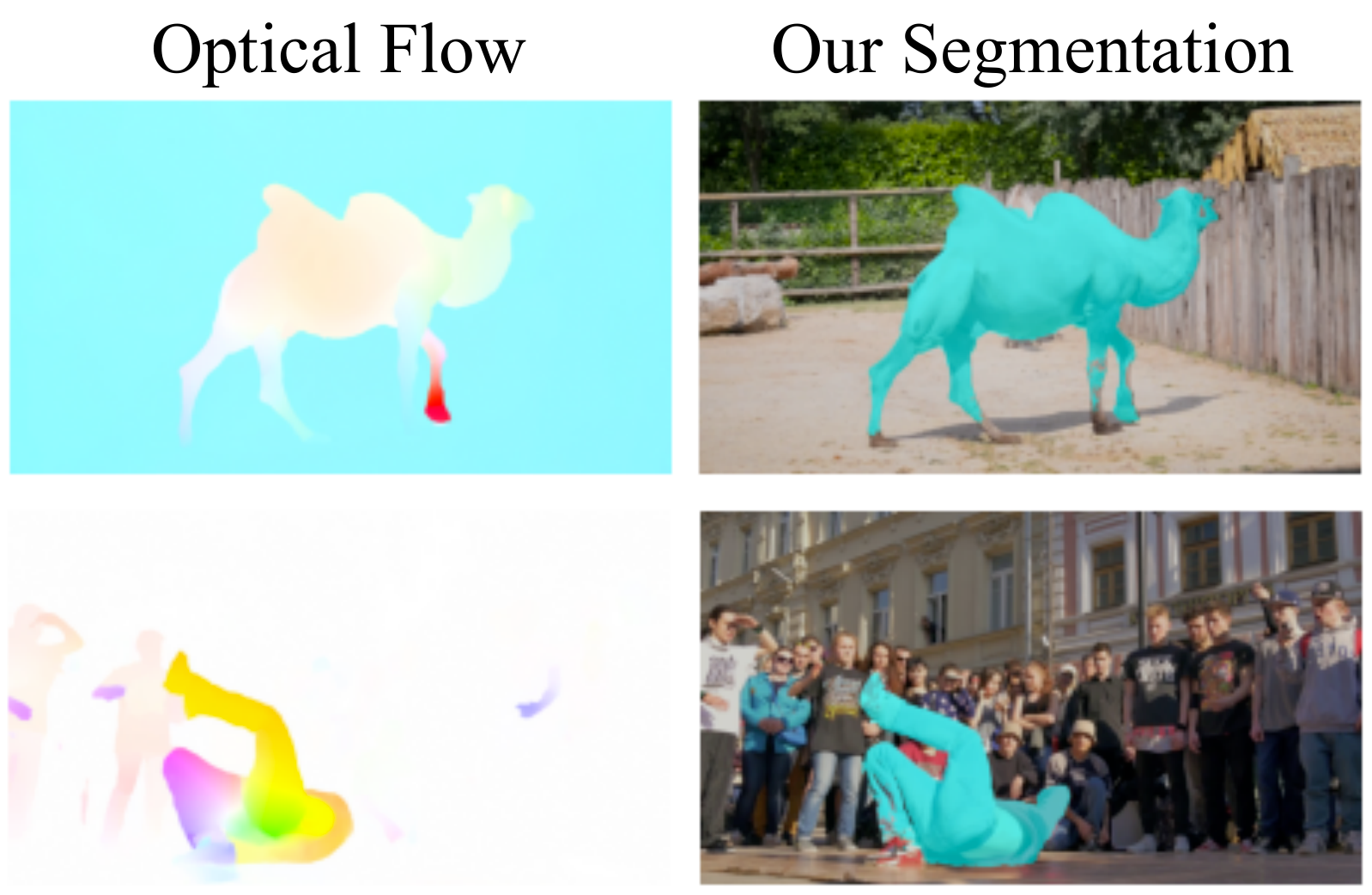}}
\vspace{5pt}
\caption{
Visualizations
\vspace{-2pt}
}
\end{subfigure}
\caption{IMAS generates high-quality segmentation and is robust to uninformative or even misleading motion signals. Best viewed in color and zoom in. More in \cref{fig:visualizations}. $\dagger$: with post-processing.\vspace{-15pt}}
\label{fig:teaser}
\end{figure}
}

\iffalse
\def\figTeaser#1{
\begin{figure}[#1]
\centering
%\vspace{-5pt}
\centerline{\includegraphics[width=1.1\linewidth]{figures/fig1.pdf}}
\vspace{-5pt}
\caption{\textbf{Our framework at training time.} Our method performs unsupervised video object segmentation with both motion and appearance supervision in an object discovery stage and an appearance-based refinement stage. Only the backbone and segmentation head are present at inference time. \vspace{-5pt}}
\label{fig:teaser}
\end{figure}
}
\fi

\section{Introduction}
\label{introduction}
\tabAdvantage{!t}
\figTeaser{!t}

Video object segmentation is a widely researched topic in the field of computer vision \cite{perazzi2016benchmark,li2013video,brox2010freiburg,ochs2013segmentation}. However, many video object segmentation algorithms require pixel-wise manual annotation that is often costly to obtain \cite{lu2019see,mahadevan2020making,zhen2020learning,li2018unsupervised,schmidt2022d2conv3d,ren2021reciprocal,zhou2020motion,ji2021full,cheng2021rethinking,cheng2022xmem,liu2022learning,miao2022region}. In recent years, popularity has been gained in the task of unsupervised video object segmentation (UVOS) \cite{yang2019unsupervised,yang2021self,meunier2022driven,liu2021emergence,lamdouar2021segmenting,xie2022segmenting,choudhury2022guess}, which learns to segment foreground objects in videos without human annotation. Most of these methods ground segmentation on motion cues. Taking optical flow as input, these methods segment out areas with consistent motion patterns. Furthermore, recent methods \cite{yang2021self, meunier2022driven,lamdouar2021segmenting,xie2022segmenting} propose to take motion as the only input for segmentation, showing the effectiveness of motion in UVOS.

Truly, motion cues, which are also leveraged by human beings, work well in identifying the main objects even in scenes with complicated background textures. However, overly grounding segmentation on motion has its own caveats: objects or object parts may be slowly moving or even static in a scene, which blends themselves with background in optical flow and causes difficulty in segmentation.

Approaches \cite{liu2021emergence,choudhury2022guess} have been proposed to tackle this problem. AMD \cite{liu2021emergence} proposed to self-supervisedly train an optical flow network jointly with a segmentation network that takes images as input. The goal is to reconstruct the next frame with a rigid motion model assuming piecewise constant flow w.r.t predicted masks\footnote{In this work, we refer to the modeling on how moving objects generate optical flow as the \textit{motion model} (\eg, constant/affine motion model), which is different from \textit{optical flow models/networks} (\eg, RAFT \cite{teed2020raft}).}. Only the segmentation network is used at inference time. Using motion as a training signal for objectness and taking images as input, AMD is able to detect non-moving objects in videos. With an affine motion model and an optical flow model from synthetic data, GWM \cite{choudhury2022guess} improves the segmentation quality further and sets new records on several UVOS tasks.

Nonetheless, these appearance-based methods still suffer from undesired properties inherent in motion cues, which are observed to mislead the model in training: \textbf{1)} objects with articulated or deformable parts may present complex motion patterns unable to be satisfactorily fitted with a hand-crafted motion model (\eg, the dancer in \cref{fig:teaser}(b)), causing under-segmentation; \textbf{2)} misleading common motion could trick the model into segmenting the object with reflections or shadows moving in the same direction/speed (\eg, the swan in \cref{fig:refine-example}), causing over-segmentation. Despite the appearance inductive bias in neural networks \cite{geirhos2018imagenettrained}, the incentive to fit the motion often overrides such property.

% Despite the success of current appearance-based UVOS methods, \textit{several caveats from leveraging motion signals still hold}:
% \textbf{1)} Articulated or deformable objects often have complex non-rigid patterns that are unable to be satisfactorily fitted with a hand-crafted motion model (e.g., piecewise constant or affine motion model in \cite{liu2021emergence, choudhury2022guess}), leading to large false negatives. 
% \textbf{2)} Reflections and shadows often follow very similar motion as the object and thus provides a noisy supervision signal for the segmentation model, resulting in false positives. Even though neural network models often has inductive bias towards appearance (e.g., texture \cite{geirhos2018imagenettrained}), the incentive to minimizes the loss easily overrides such property empirically. 

Inspired by humans that leverage motion cues for foreground object discovery in a complicated scene and then appearance cues for more precise segmentation, we propose a UVOS framework with two training stages: \textbf{1)} an object discovery stage that learns objectness from motion supervision; \textbf{2)} a subsequent refinement stage that uses appearance supervision to correct the model's misconceptions learned from misleading motion cues in the previous stage.

% \textbf{1)} an object discovery stage that leverages motion supervision flexibly through residual pathway; \textbf{2)} an appearance-based refinement stage with both low- and high-level appearance supervision to correct misconceptions learned from misleading motion cues.

In the object discovery stage, we devise a flexible motion model with a learnable residual pathway to allow \textit{implicit} appearance learning when motion cues conflict with appearance. In the refinement stage, we propose to use appearance cues as \textit{explicit} supervision to correct misconceptions from motion supervision. The two stages lead to \underline{I}mproved UVOS with \underline{M}otion-\underline{A}ppearance \underline{S}ynergy (\textbf{IMAS}).

% Although difficult for current methods, we observe that the above cases are solvable by human beings who acquire the ability to localize and segment foreground objects through motion and appearance signals. 

% Instead of adapting the optical flow signals, we \textit{explicitly introduce appearance cues as supervision} so that the segmentation learning occurs in two steps: \textbf{1)} With a flexible learned motion model, motion is first used for localization and coarse segmentation of the foreground objects. \textbf{2)} Both low-level and high-level appearance cues are leveraged as training signals for segmentation correction and refinement to segment out the complete foreground objects. Neither of the two steps requires human supervision, which makes our method fully unsupervised.

In addition, we propose to also use motion-appearance relationship as an unsupervised hyperparam tuner. By using motion-semantic alignment as a proxy for segmentation quality, we could tune the hyperparameters without manual annotations. We demonstrate its effectiveness on critical hyperparams previously tuned with human annotation or hand-crafted hyperparam-specific metrics. This technique is \textit{model-agnostic} and applicable to other UVOS methods.

IMAS demonstrates performance superior to previous work on various common UVOS benchmarks without heavyweight backbones, non-standard layers, or significant post-processing. For example, our method improves on previous methods by over $8.3\%$ ($5.6\%$) with (without) post-processing on DAVIS16\cite{perazzi2016benchmark} benchmark (\cref{fig:teaser}).

\textbf{Our work has the following contributions}:
\ol{enumerate}{0}{
\item We devise a learnable residual pathway to resolve cue conflicts for improved implicit appearance learning.
\item We propose explicit appearance supervision to correct misconceptions learned from misleading motion cues.
\item We put forward motion-semantic alignment as a model-agnostic annotation-free hyperparam tuner. % and demonstrate effectiveness on two critical hyperparams in motion learning
\item We perform extensive evaluations and show significant improvements in UVOS tasks in challenging scenarios.
}

We intend to release our code for future research and applications.

\section{Related Work}
\label{related_work}
\textbf{Unsupervised video object segmentation} (UVOS) requires segmenting prominent objects from video sequences without human annotation. Mainstream benchmarks on UVOS \cite{perazzi2016benchmark,li2013video,brox2010freiburg,ochs2013segmentation} define the task as a figure-ground problem with binary predictions in which salient objects are the foreground. Despite the name, several previous UVOS methods require \textit{supervised} \mbox{(pre-)training} on large-scale images or videos with manual annotation \cite{lu2019see,koh2017primary,faktor2014video,zhen2020learning,li2018unsupervised,yang2021dystab,ren2021reciprocal,zhou2020motion}. In contrast, we focus on a line of work named \textit{fully unsupervised VOS}, which does \textit{not} rely on any human annotation at either \textit{training or inference} time. For fairness, we compare with previous literature on fully UVOS.

\textbf{Motion segmentation} aims at exploiting the motion information to separate foreground and background, typically through using optical flow from a pretrained model.
% NLC \cite{faktor2014video} and ARP \cite{koh2017primary} perform temporal clustering for motion segmentation but leverages supervised saliency models. 
FTS \cite{papazoglou2013fast} utilizes motion boundaries for segmentation. SAGE \cite{wang2017saliency} additionally considers edges and saliency priors jointly with motion. CIS \cite{yang2019unsupervised} uses independence between foreground and background motion as the goal for foreground segmentation. However, such property is not always satisfied in complex real-world motion patterns. MG \cite{yang2021self} leverages attention mechanisms to group pixels with similar motion patterns. SIMO \cite{lamdouar2021segmenting} and OCLR \cite{xie2022segmenting} propose to generate synthetic data for segmentation supervision, with the latter supporting individual segmentation on multiple objects, but both methods rely on human-annotated sprites for realistic shapes in artificial data synthesis. These motion segmentation methods leverage optical flow as input and thus suffer when foreground objects are moving at a similar speed to the background.

Recently, \textbf{appearance-based UVOS} methods found a way around by using motion only as a supervisory signal for learning the moving tendency. These methods use raw images as the input to the segmentation model and thus could detect objects without relative motion. AMD \cite{liu2021emergence} proposes to jointly learn a segmentation model and an optical flow model so that flow prediction is constructed to be piecewise constant according to segmentation. The flow prediction is then supervised by minimizing the reconstruction loss from warped image guided by it. GWM \cite{choudhury2022guess} proposes to match the piecewise affine flow with filtered optical flow generated by a pretrained flow model \cite{teed2020raft} and is the current state-of-the-art of several UVOS tasks.

Despite the success of appearance-based UVOS, \textit{several caveats from motion signals still hold}:
\textbf{1)} Articulated or deformable objects often have complex non-rigid motion hard to fit into hand-crafted motion models, causing false negatives. 
\textbf{2)} Naturally-occurred common motion (\eg, reflections and shadows) often provides misleading supervision, resulting in false positives. To tackle such issues, we use motion cues to discover the foreground objects with a learnable residual pathway that allows implicit appearance learning. Subsequently, appearance cues are explicitly leveraged as training signals for misconception correction.

\def\tabMainComparison#1{
\begin{table}[#1]
\setlength{\tabcolsep}{2.5pt}
\centering
\begin{tabular}{l||c@{}ccc}
\shline
Methods & Post-process & \textbf{DAVIS16} & \textbf{STv2} & \textbf{FBMS59} \\
\shline
% \textit{Without post-processing:} & & & \\
SAGE \cite{wang2017saliency}          & & 42.6 & 57.6 & 61.2 \\ % CIS
% \gray{NLC$^**$}     & & \gray{55.1} & \gray{67.2} & \gray{51.5} \\ % CIS, edge detection supervision
CUT \cite{keuper2015motion}           & & 55.2 & 54.3 & 57.2 \\ % CIS
FTS \cite{papazoglou2013fast}         & & 55.8 & 47.8 & 47.7 \\ % CIS
EM \cite{meunier2022driven}           & & 69.8 & --   & --   \\ % GWM
% \gray{ARP$^**$} \cite{koh2017primary} & & \gray{76.2} & \gray{57.2} & \gray{59.8} \\ % CIS, saliency supervision
CIS \cite{yang2019unsupervised}       & & 59.2 & 45.6 & 36.8 \\
MG \cite{yang2021self}                & & 68.3 & 58.6 & 53.1 \\
AMD \cite{liu2021emergence}           & & 57.8 & 57.0 & 47.5 \\
SIMO \cite{lamdouar2021segmenting}    & & 67.8 & 62.0 & -- \\
GWM \cite{choudhury2022guess}         & & 71.2 & 66.7 & 60.9 \\
\gray{GWM$^*$ \cite{choudhury2022guess}} & & \gray{71.2} & \gray{69.0} & \gray{66.9} \\
OCLR$^\dagger$ \cite{xie2022segmenting}  & & 72.1 & 67.6 & 65.4 \\
\hline
% IMAS (stage 1) & & 76.1 & 67.2 & 64.5 \\
\textbf{IMAS-LR}    & & 77.3 & 67.9 & 67.1 \\
\textbf{IMAS}      & & \textbf{77.7} & \textbf{68.6} & \textbf{67.4} \\
                   & & \textcolor{Green}{\bf\small (+5.6)} & \textcolor{Green}{\bf\small (+1.0)} & \textcolor{Green}{\bf\small (+2.0)} \\
\shline
% \textit{With post-processing:} & & & \\
% TODO: earlier methods from AMD paper.
CIS \cite{yang2019unsupervised}                & CRF + SP$^\ddagger$                & 71.5        & 62.0        & 63.6 \\
\gray{GWM$^*$} \cite{choudhury2022guess}       & \gray{CRF + SP$^\ddagger$} & \gray{73.4} & \gray{72.0} & \gray{68.6} \\
\gray{OCLR$^\dagger$} \cite{xie2022segmenting} & \gray{Tuning DINO$^\ddagger$} 
                                 & \gray{78.9} & \gray{71.6} & \gray{68.7} \\
\hline
% IMAS (stage 1) & 78.4 & 71.4 & 66.7 \\
\textbf{IMAS-LR} & CRF only & 79.5 & 72.2 & 68.9 \\
\textbf{IMAS}   & CRF only & \textbf{79.8} & \textbf{72.7} & \textbf{69.0} \\
                &          & \textcolor{Green}{\bf\small (+8.3)} & \textcolor{Green}{\bf\small (+10.7)} & \textcolor{Green}{\bf\small (+5.4)} \\
\shline
\end{tabular}
% We outperform methods that use complex architecture, with simple architecture.
\caption{\textbf{Our method achieves significant improvements over previous methods on common UVOS benchmarks}. \textbf{IMAS-LR} indicates low-level refinement only (no $f_\text{aux}$ used).
%$*$: NLC and ARP uses edge detection and saliency supervision, respectively.
$*$: uses heavier architecture (Swin-Transformer \cite{liu2021swin} with MaskFormer \cite{cheng2021maskformer} segmentation head) orthogonal to VOS method and thus is not a fair comparison with us.
$\dagger$ leverages manually annotated shapes from large-scale Youtube-VOS \cite{xu2018youtube} to generate synthetic data. 
$\ddagger$: \textit{SP}: significant post-processing (\eg, multi-step flow, multi-crop ensemble, and temporal smoothing). \textit{Tuning DINO}: performing contrastive learning on a pretrained DINO ViT model \cite{caron2021emerging,dosovitskiy2020image} as post-processing and thus is not a fair comparison with us.
Our post-processing is a \textit{CRF pass only} with an off-the-shelf CRF library.
\vspace{-15pt}}
% TODO: CIS is from AMD (add the description in text).
% TODO: uses Swin-Transformer and MaskFeat which is orthogonal to segmentation method.
% DINO finetuning is orthogonal to our method.
\label{tab:main-comparison}

\end{table}
}

\def\figMethod#1{
\begin{figure*}[#1]
%\begin{subfigure}{\linewidth}
\centering
%\vspace{-4pt}
\centerline{\includegraphics[width=1.05\linewidth]{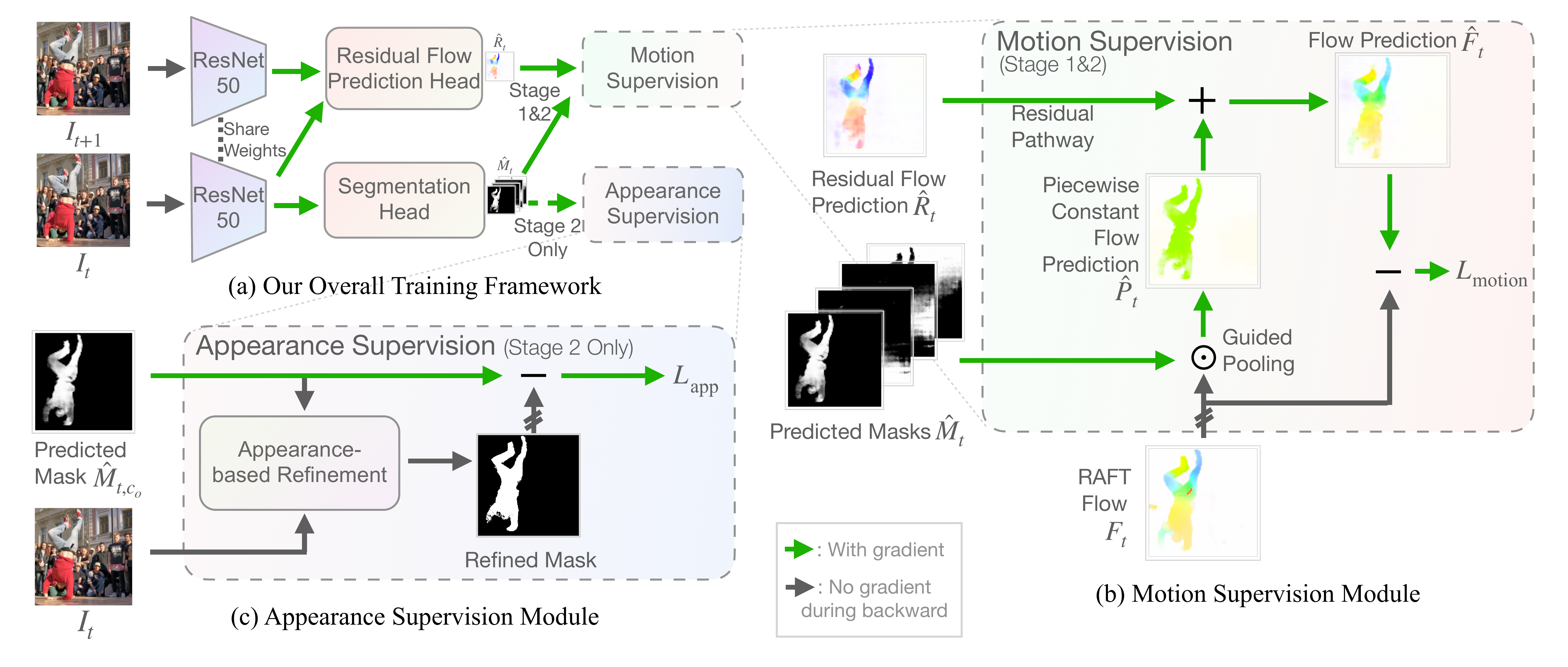}}
\vspace{-4pt}
%\end{subfigure}
\caption{Our method performs unsupervised video object segmentation with an object discovery stage and an appearance-based refinement stage. 
\textbf{(a)~Our framework at training time}. Only the backbone and segmentation head are present at inference time, requiring only image input. \textbf{(b)~Our motion supervision module} aims at reconstructing the reference RAFT flow with a piecewise constant flow pathway constructed from the predicted masks and a pixel-wise residual flow prediction pathway. The residual pathway relieves the model from strictly complying with the noisy and non-uniform flow, which allows implicit appearance learning when motion supervision conflicts with appearance knowledge. \textbf{(c)~Our appearance supervision module} further corrects the misconception learned from motion supervision. This module supervises the predicted mask by a refined version of the mask based on both low- and high-level appearance.}
\label{fig:method}
\end{figure*}
}

\iffalse
\def\figMotionSup#1{
\begin{figure}[#1]
\begin{subfigure}{\linewidth}
  \centering
  %\vspace{-4pt}
  \centerline{\includegraphics[width=1.0\linewidth]{figures/fig2.pdf}}
  \vspace{-4pt}
\end{subfigure}
\caption{Our \textbf{motion supervision module} aims at reconstructing the flow with a piecewise constant pathway constructed from the predicted masks and a pixel-wise residual flow prediction. The residual pathway implicitly allows the model to capture more appearance information without strictly complying with the noisy flow in a piecewise constant manner.}
\label{fig:motion-sup}
\end{figure}
}

\def\figAppearanceSup#1{
\begin{figure}[#1]
\begin{subfigure}{\linewidth}
  \centering
  %\vspace{-4pt}
  \centerline{\includegraphics[width=1.0\linewidth]{figures/fig3.pdf}}
  \vspace{-4pt}
\end{subfigure}
\caption{Our \textbf{appearance supervision module} aims at correcting the misconception learned from motion in the object discovery stage. This module supervises the predicted mask by a refined version of the mask based on both low-level and high-level appearance information.}
\label{fig:appearance-sup}
\end{figure}
}
\fi

\def\figRefineExample#1{
\begin{figure}[#1]
\vspace{-15pt}
\centerline{\includegraphics[width=1.1\linewidth]{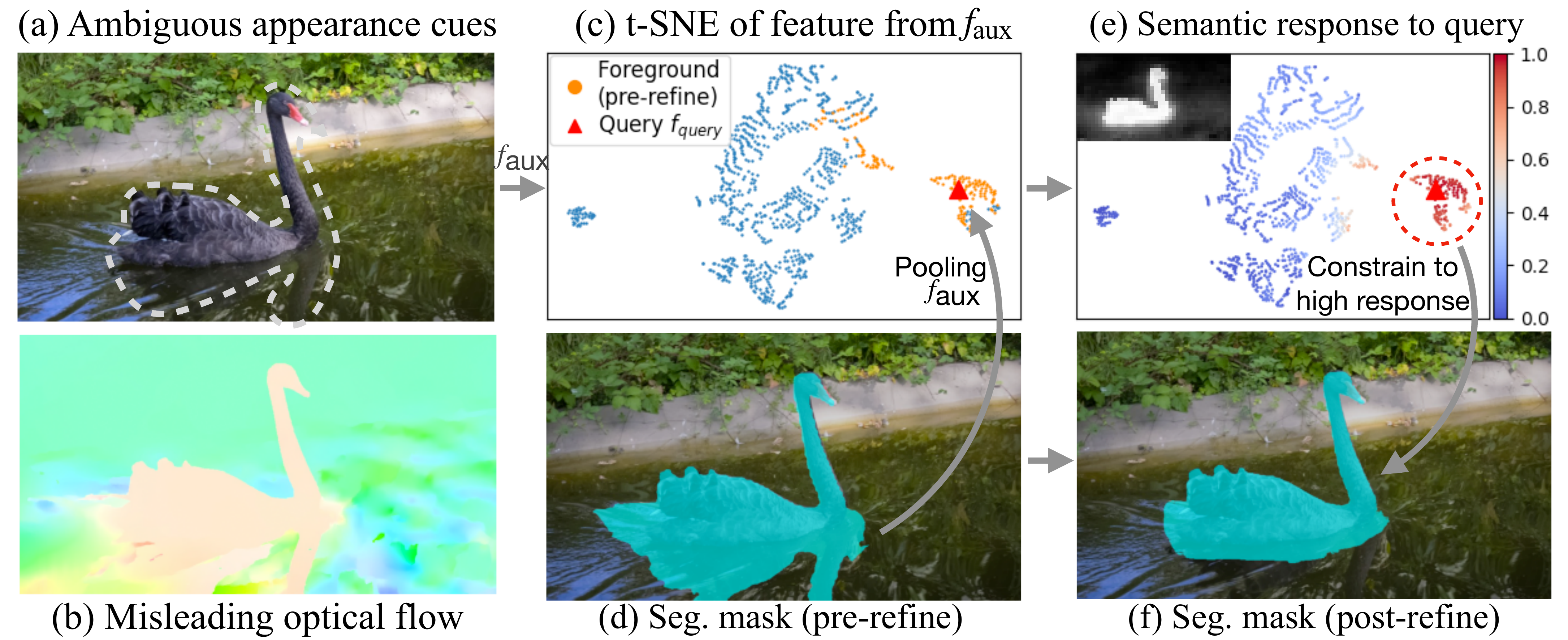}}
\vspace{-5pt}
\caption{\textbf{Semantic constraint mitigates false positives from naturally-occurring misleading motion signals.} The reflection has semantics distinct from the main object and is thus filtered out. The refined mask is then used as supervision to disperse the misconception in stage 2. Best viewed in color and zoom in. \vspace{-5pt}}
\label{fig:refine-example}
\end{figure}
}

\def\figTuningExample#1{
\begin{figure}[#1]
\vspace{-15pt}
\centerline{\includegraphics[width=1.0\linewidth]{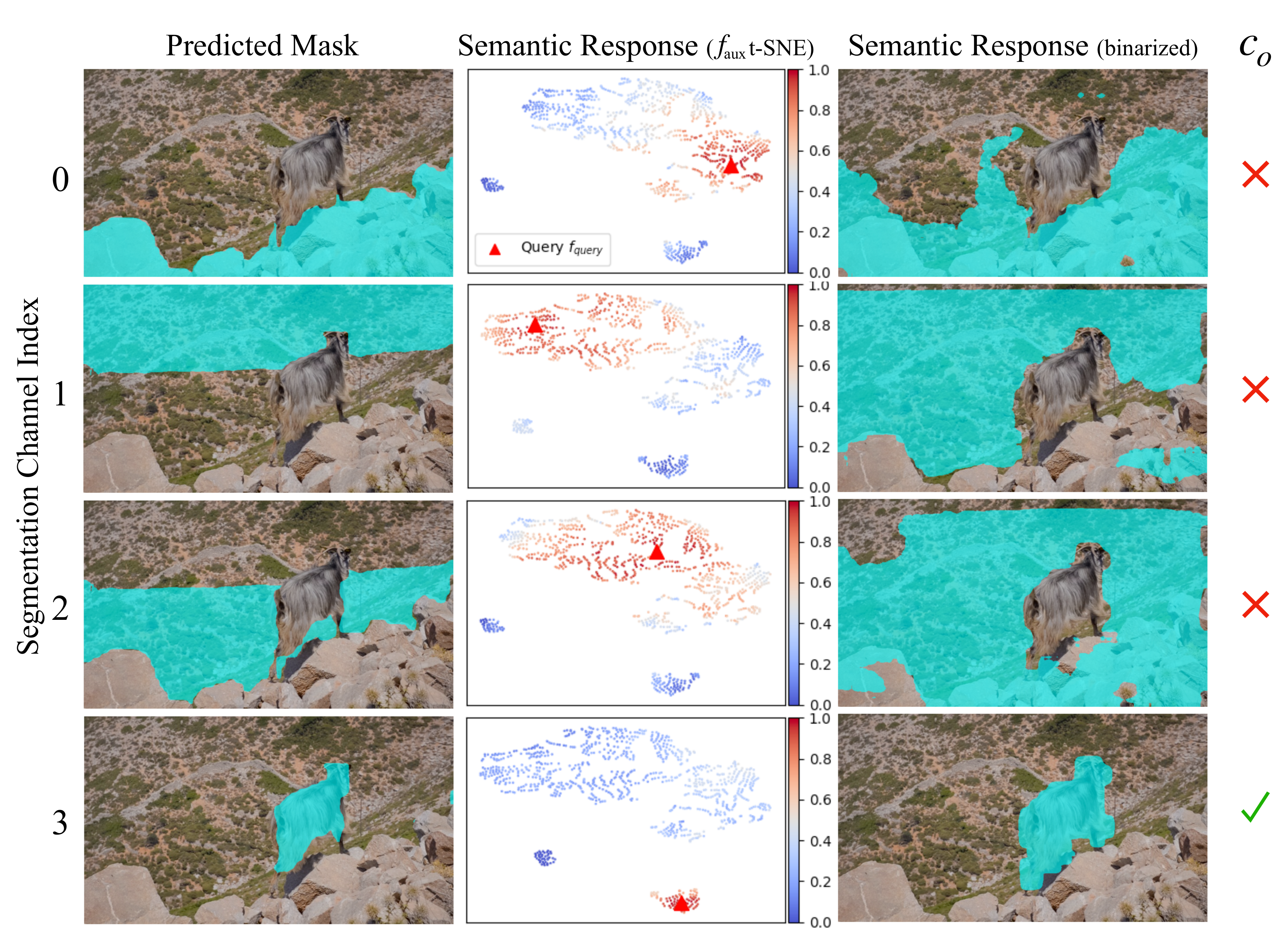}}
\vspace{-10pt}
\caption{\textbf{Selecting objectness channel $c_o$ with motion-semantic alignment.} Only channel $3$ gives a high IoU between the mask prediction $\hat M_{t,c}$ and the semantic response $S_{t,c}$. If this holds true in most frames (quantified by mean IoU), we set $c_o$ as $3$.\vspace{-10pt}}
\label{fig:tuning-example}
\end{figure}
}

\figMethod{!t}
\section{Method}
\label{method}

As in \cref{fig:method}, IMAS consists of a motion-supervised object discovery stage and a subsequent appearance-supervised refinement stage. 
The former uses a flexible motion model with a learnable residual pathway to allow implicit appearance learning when motion cues conflict with appearance. The latter explicitly uses low- and high-level appearance supervision to correct misconceptions learned from misleading motion.
%The former one uses a flexible motion model with learnable residual pathway for implicit appearance learning, and the latter one learns to refine the segmentation with appearance supervision to correct misconceptions learned from motion.
Neither stage requires human annotation, which makes IMAS \textit{fully unsupervised}.

% \textbf{1)} an object discovery stage that leverages motion supervision flexibly through residual pathway; \textbf{2)} an appearance-based refinement stage with both low- and high-level appearance supervision to correct misconceptions learned from misleading motion cues.

% \textbf{1)} Through a flexible motion model with residual pathway for implicit appearance learning, motion signals are first used for localization and coarse segmentation of the foreground objects. \textbf{2)} Both low-level and high-level appearance cues are leveraged as training signals for segmentation correction and refinement to segment out the complete foreground objects. 

We present the problem setting in \cref{sec:method_problem_setting} and describe our contributions in the first/second stage in \cref{sec:method_motion_supervision}/\cref{sec:method_refinement}, respectively. In \cref{sec:method_tuning}, we present motion-semantic alignment as a model-agnostic unsupervised hyperparam tuner.

% Need to stress that our motion model is learned.

% Instead of adapting the optical flow signals, we \textit{explicitly introduce appearance cues as supervision} so that the segmentation learning occurs in two steps: \textbf{1)} With a flexible learned motion model, motion is first used for localization and coarse segmentation of the foreground objects. \textbf{2)} Both low-level and high-level appearance cues are leveraged as training signals for segmentation correction and refinement to segment out the complete foreground objects. Neither of the two steps requires human supervision, which makes our method fully unsupervised.

\subsection{Problem Setting}
\label{sec:method_problem_setting}

Let $I_t \in \mathbb{R}^{3 \times h \times w}$ be the $t^{\text{th}}$ frame from a sequence of $T$ RGB frames, where $h$ and $w$ are the height and width of the image, respectively. The objective of UVOS is a binary segmentation mask $M_t \in \{0, 1\}^{h \times w}$ for each timestep $t$, with $1$ indicating foreground and $0$ otherwise. To evaluate a method on UVOS, we compute the mean Jaccard index $\mathcal{J}$ (\ie, mean IoU) between predicted segmentation mask $M_t$ and ground truth $G_t$. Since we are performing fully unsupervised video object segmentation, ground truth mask $G_t$ is assumed to be unavailable and no human-annotated data are used throughout training and inference.

\subsection{\fontsize{10.5}{10.5}\selectfont\bf\mbox{Object Discovery with Flexible Motion Supervision}}
\label{sec:method_motion_supervision}

Following appearance-based UVOS work\cite{liu2021emergence,choudhury2022guess}, we use motion to supervise foreground objectness learning. However, rather than hand-crafted motion models (\eg, piecewise constant or affine), we propose a \textit{learnable} motion model with a residual pathway to facilitate implicit appearance learning when motion is uninformative or conflicts with appearance (\cref{fig:method}(b)).

Let $f(I_t) \in \mathbb{R}^{K \times H \times W}$ be the feature of $I_t$ extracted from a ResNet-50 \cite{he2016deep} backbone $f(\cdot)$, where $K$, $H$, and $W$ are the number of channels, height, and width of the feature. Let $\hat M_t = g(f(I_t)) \in \mathbb{R}^{C \times H \times W}$ be $C$ soft segmentation masks extracted with segmentation head $g(\cdot)$, where $g(\cdot)$ is a lightweight fully convolutional module composed of three Conv-BN-ReLU\cite{ioffe2015batch} layers. \Verb/Softmax/ is taken across channel dimension inside $g(\cdot)$ so that the $C$ soft masks sum up to 1 for each of the $H \times W$ positions. As in \cite{liu2021emergence}, although there are $C$ segmentation masks competing for each pixel (i.e., $C$ output channels in $\hat M_t$), \textit{only one of them} corresponds to the foreground, with the rest capturing different background patches. We define $c_o$ as the object channel index, whose value is obtained in \cref{sec:method_tuning}.

Following \cite{yang2021self,lamdouar2021segmenting,xie2022segmenting,choudhury2022guess}, we use off-the-shelf optical flow model RAFT \cite{teed2020raft} trained on synthetic datasets\cite{dosovitskiy2015flownet,mayer2016large} without human supervision to provide motion cues between consecutive frames. Let $F_t \in \mathbb{R}^{2 \times H \times W}$ be the optical flow output from RAFT \cite{teed2020raft} from timestep $t$ to $t+1$.

% \figMotionSup{!t}
\noindent\textbf{Piecewise constant pathway.} We first pool the flow according to each mask to form $C$ flow vectors $\hat P_{t,c} \in \mathbb{R}^{2}$:
\vspace{-5pt}
\begin{equation}
\hat P_{t, c} = \phi_2(\text{GuidedPool}(\phi_1(F_t), \hat M_{t, c})) 
\end{equation}
where $\text{GuidedPool}(F, M) = \frac{\sum_{p=1}^{HW} (F \odot M)[p]}{\sum_{p=1}^{HW}M[p]}$, with $[p]$ spatial indexing at position $p$ and $\odot$ element-wise multiplication. Following \cite{liu2021emergence}, $\phi_1$ and $\phi_2$ are two-layer lightweight MLPs with $64$ hidden units that transform each of the motion vectors independently before and after pooling, respectively.
We then construct a flow prediction $\hat P_t \in \mathbb{R}^{2 \times H \times W}$ according to the soft segmentation mask:
\vspace{-5pt}
\begin{equation}
\hat P_t = \sum_{c=1}^{C} \text{Broadcast}(\hat P_{t, c}, \hat M_{t, c}) 
\end{equation}
where $\text{Broadcast}(\hat P_{t, c}, \hat M_{t, c})[p] = \hat P_{t, c} \odot (\hat M_{t, c}[p])$.

With this pathway, as the soft segmentation approaches binary from random during training, flow prediction approaches a piecewise constant function w.r.t each segmentation mask, which captures the common fate in visual grouping\footnote{Since the ultimate goal is to learn a binary mask with a piecewise constant flow, we name the motion model piecewise constant motion model.}. Previous methods either directly supervise $\hat P_t$ with image warping loss for self-supervised learning \cite{liu2021emergence} or matches $\hat P_t$ and $F_t$ by minimizing the discrepancies up to an affine factor (\ie, up to first order) \cite{choudhury2022guess}.

Nonetheless, such hand-crafted non-learnable motion models underfit the complex motion patterns in the real world. We observe that when a deformable or articulated object is present in the scene, the segmentation model often separates part of the object into different channels to minimize the loss, despite similar color or texture suggesting otherwise. This causes incomplete foreground. \cite{choudhury2022guess} proposes to relieve the effect by using only 2 mask channels. However, this further reduces the fitting power and renders the inability to capture scenes with complex background.

% Note: goat with flow if asked to provide example

\noindent\textbf{Learnable residual pathway.} 
Although an object may exhibit complex non-rigid motion in the real world, the motion of each object part is often relative to the main motion of the object (\eg, in \cref{fig:method}(b), the dancer's feet are moving relative to the dancer's body). Similarly, many motion patterns, such as rotation or depth changes, can be modeled by the residual flow.
% Although motion pattern of foreground objects may be complicated in the real world, we observe that the motion of parts of an object is often relative to the main motion of the object (e.g. the dancer's feet are moving relative to the dancer's body). 
Therefore, rather than more complicated \textit{hand-crafted} motion models (e.g., quadratic), %we propose a \textit{learnable} residual pathway with non-piecewise constant relative motion in mind.
we add a \textit{learnable} residual pathway $\hat R_t$ in addition to piecewise constant pathway $\hat P_t$ to form the final flow prediction $\hat F_t$. % When non-uniform motion signals conflict with uniform or previously-seen appearance cues, $\hat R_t$ could relieve $\hat P_t$ from fitting complex motion so that implicit appearance knowledge can be learned and preserved.
% $\hat R_t$ is from a shallow fully convolutional module learned by back-propagation, with a scaled $\tanh(\cdot)$ activation at the output to constrain the predicted residuals (relative motion) within small values so that the network could not fit the flow of the object directly.

Let $h(\cdot)$ be a lightweight module with three Conv-BN-ReLU blocks that take the concatenated feature of a pair of frames $\{I_t, I_{t+1}\}$ as input and predicts $\hat R_t' \in \mathbb{R}^{C \times 2 \times H \times W}$, which includes $C$ flows with per-pixel upper bound $\lambda$:
\vspace{-5pt}
\begin{align}
\hat R_t' = \lambda\tanh(h(\text{concat}(f(I_t), f(I_{t+1})))
\end{align}
where upper bound $\lambda$ is set to $10$ pixels unless stated otherwise.
The $C$ flows are then composed to form the final residual flow $\hat R_t$ by the mask predictions and summed up with the piecewise constant pathway, forming the final flow prediction $\hat F_t$:
\vspace{-5pt}
\begin{align}
\hat R_t &= \sum_{c=1}^{C} \hat R_{t, c}' \odot \hat M_{t, c} \\
\hat F_t &= \hat P_t + \hat R_t
\end{align}
In this way, $\hat F_t$ additionally takes into account relative motion that is within $(-\lambda, \lambda)$ for each spatial location. Rather than being asked to exactly obey hand-crafted motion rules from the often imprecise motion signal, the added residual pathway allows the neural network to \textit{implicitly} preserve its knowledge and decision from the appearance side when the motion supervision is uninformative or conflicts with appearance as input, leading to better segmentation results. % the segmentation network is allowed to leverage implicit inductive bias for appearance learning and is encouraged to predict the segmentation that is easiest to fit from appearance (\eg, colors and textures).

% In this way, when the motion signal is misleading or even conflicts with the appearance, rather than forcing the model to learn a segmentation that optimizes piecewise constant flow to match the often imprecise signals from motion, the added residual pathway allows the neural network to \textit{implicitly} preserve its decision from the appearance side, leading to better segmentation results.

The training objective of stage 1 is to minimize the L1 loss between the predicted reconstruction flow $\hat F_t$ and target flow $F_t$ to learn segmentation by predicting the correct flow:
\vspace{-5pt}
\begin{equation}
L_{\text{stage 1}} = L_{\text{motion}} = \frac{1}{HW}\sum_{p=1}^{HW}||\hat F_t[p] - F_t[p]||_1
\end{equation}
\vspace{-10pt}
%

% Note: Visualization: neighboring patches with similar texture and similar activations are supervised with different motion directions or speed

\subsection{\mbox{Refinement with Explicit Appearance Supervision}}
\label{sec:method_refinement}

% \figAppearanceSup{!t}

Even though allowing implicit appearance learning in stage~1 greatly improves the segmentation quality, we still observe unsatisfactory segmentation when the foreground is moving in a fast yet unpredictable pattern. In such cases, the lightweight residual prediction module may mispredict the residual for some patches. This causes the segmentation model to assign some probabilities to multiple predicted masks, which blends motion from multiple channels to match the target flow. This manifests itself in unconfident predictions that often become false negatives when we take \Verb/argmax/ across channels to get binary masks $M_t$.

% In additional to the implicit appearance learning in stage~1, we propose to explicitly use both low- and high-level appearance cues as supervision to correct the misconception from misleading motion cues.
% For example, the under part of the non-moving leg has motion vectors much more similar to the ground instead of the body that is moving. We observe that the model produces unconfident predictions for the lower part of the leg due to its optimality in blending in the flow from both the body and the ground.

One naive option is to penalize unconfident predictions with entropy regularization. However, we found that such regularization, when strong enough to relieve the effect, disturbs learning and makes the neural network scatter the foreground object into several parts in different channels. %However, we found that training becomes unstable when such regularization is strong enough to relieve such an effect, where the animal is broken into several parts, each with low motion variation.

Although optimal in terms of flow prediction, such segmentation is often suboptimal from an appearance perspective and unnatural to human beings who leverage both motion and appearance signals in figure-ground separation.
For example, the segmentation prediction $\hat M_{t, c_o}$ in \cref{fig:method}(c) ignores part of the dancer's leg, despite the ignored part sharing very similar color and texture with the included parts.
%The segmentation in camel does not fit the whole animal, despite no boundaries (i.e., edges) present in any part of the leg.

% Inspired by this, instead of engineering motion signals to reduce ambiguity, we propose to train a model with misconceptions corrected through explicitly leveraging both low-level and high-level appearance signals in images.

Inspired by this, we propose to train a model with misconceptions corrected by \textit{explicitly} leveraging both low- and high-level appearance signals in images.

% Simplify
% We propose to use training-free techniques that refine the masks based on appearance. Balancing implementation complexity and refinement quality, we select fully-connected conditional random field, for which highly optimized implementations exist, with other refinement methods (e.g. \cite{barron2016fast}) left for future investigations.
\noindent\textbf{Appearance supervision with low-level cues.} With the model learned from motion signals in stage 1, we obtain the mask prediction $\hat M_{t, c_o}$ of $I_t$, where $c_o$ is the objectness channel that could be found without annotation (\cref{sec:method_tuning}). We then apply a training-free technique that refines the masks based on low-level appearance.

Balancing implementation complexity and refinement quality, we select fully-connected conditional random field~(CRF) \cite{krahenbuhl2011efficient}, a technique that refines the value of each prediction based on other pixels with an appearance and a smoothness kernel, for which highly optimized implementations exist. 
% Then, we apply fully-connected conditional random field (CRF) \cite{krahenbuhl2011efficient}, a technique that refines the value of each pixel based on other pixels with an appearance and a smoothness kernel, to propagate the segmentation mask with both appearance and relative position in mind. 
% With appearance awareness, the refined mask recognizes that despite the lower part of the legs having different motion as body and similar motion as ground, legs belong to the body due to similar appearance.
The refined masks $\hat M'_{t, c_o}$ are then used as supervision to provide explicit appearance signals in training:
\vspace{-5pt}
\begin{align}
\hat M'_{t, c_o} &= \text{CRF}(\hat M_{t, c_o}) \label{eq:crf_refinement}\\[-2pt]
L_{\text{app}} &= \frac{1}{HW} \sum_{p=1}^{HW}||\hat M_{t, c_o}[p] - \hat M'_{t, c_o}[p]||_2^2
\end{align}
\vspace{-12pt}

\noindent Since stage 2 is mainly misconception correction and thus much shorter than stage 1, we generate the refined masks for supervision only once between the two stages for efficiency.

The total loss in stage 2 is a weighted sum of both motion and appearance loss:
\vspace{-5pt}
\begin{equation}
L_\text{stage 2} = w_\text{app} L_\text{app} + w_\text{motion} L_\text{motion} 
\end{equation}
\vspace{-18pt}

\noindent where $w_\text{app}=2$ and $w_\text{motion}=0.1$ are balancing weights.

Attentive readers may find that CRF is also used in post-processing in previous works to upsample the predicted segmentation masks, as the predicted masks often have lower resolution than both the image and the ground truth \cite{yang2019unsupervised,choudhury2022guess} (i.e., $H \ll h$). However, the CRF in our method for appearance supervision is \textit{orthogonal} to the CRF in post-processing, as the refined masks $\hat M'_{t, c_o}$ have the same size as the prediction $\hat M_{t, c_o}$, both of which are much smaller than the image. Furthermore, the post-processing CRF, when applied to IMAS, leads to \textit{similar gains} no matter whether the refinement stage is present, which confirms orthogonality (\cref{sec:ablation}). We also admit other potentially better-performing refinement or interpolation options (\eg, \cite{banterle2012low,barron2016fast,iscen2019label,vernaza2017learning,shi2000normalized}) and leave them for future investigations.
% We also admit other better-performing options for better appearance-based supervision. %Furthermore, CRF is \textit{only an example} of the training-free techniques that we could apply to refine the mask, with other more recent techniques (e.g. \cite{barron2016fast}) also applicable to our framework.

\figRefineExample{!t}
\noindent\textbf{Appearance supervision with semantic constraint.} % With the flexible motion model and low-level appearance signals, our method is able to greatly improve upon previous methods. 
Low-level appearance is still insufficient to address misleading motion signals from naturally-occurring confounders with similar motion patterns. For example, the reflections share similar motion as the swan in \cref{fig:refine-example}, which is confirmed by low-level appearance. However, humans could recognize that the swan and the reflection have distinct semantics, with the reflections' semantics much closer to the background. % While the swan belongs to the foreground with its distinctive semantics, the reflections have semantics closer to the rest of the background.

Inspired by this, we incorporate the feature map $f_\text{aux}(I_t)$ from a frozen auxiliary ResNet\cite{he2016deep} $f_\text{aux}(\cdot)$ trained with pixel-wise self-supervised learning \cite{wang2021dense} on ImageNet \cite{ILSVRC15} without human annotation to create a semantic constraint for mask prediction  (\cref{fig:refine-example}). We first summarize foreground semantics by pooling $f_\text{aux}(I_t)$ according to mask $\hat M'_{t, c_o}$:
\vspace{-5pt}
\begin{align}
f_\text{query} &= \text{GuidedPool}(f_\text{aux}(I_t), \hat M'_{t, c_o})
\end{align}
We then obtain the semantic constraint mask $S_{t, c_o}$: 
\vspace{-5pt}
\begin{align}
S_{t, c_o} &= \text{Dilation}(\text{Threshold}(\sigma(f_\text{aux}(I_t), f_\text{query})))
\end{align}
where $\sigma(\cdot, \cdot)$ the cosine similarity, $\text{Threshold}(\cdot)$ a thresholding function with threshold value $0.3$.
Empirically, we observe that the threshold value does not require tuning because 1) having a semantic constraint that is loose often does not hurt segmentation; 2) the objects typically have very different feature responses than the background. $\text{Dilation}(\cdot)$ is the dilation operation to loosen the constraint by having a larger mask, as ResNet feature $f_\text{aux}(I_t)$ has low resolution and is not precise on the boundaries.

With semantic constraint, \cref{eq:crf_refinement} changes to:
\vspace{-5pt}
\begin{equation}
\hat M'_{t, c_o} = \text{CRF}(\hat M_{t, c_o}) \odot S_{t, c_o} 
\end{equation}
% with $S_{t, c_o}$ to ensure the refined mask is only capturing the area with relevant semantics to $f_\text{query}$.

Since semantic constraint introduces an additional frozen model $f_\text{aux}(\cdot)$ and is \textit{not} a major component for our performance gain, we benchmark both \textit{with} and \textit{without} semantic constraint for a fair comparison with previous methods, using \textbf{IMAS-LR} to denote IMAS with low-level refinement only (\ie, without semantic constraint). Our method is still fully unsupervised (\ie, without human annotation) even with semantic constraint.

% Note: It's possible that DINO offers high resolution semantic map, but due to the additional complexity we opt to our current architecture.

\figTuningExample{!t}
\subsection{\fontsize{10.5}{10.5}\selectfont\bf\mbox{Motion-Semantic Alignment as a Hyperparam Tuner}}
\label{sec:method_tuning}

Our method follows previous appearance-based UVOS work and thus has several tunable hyperparams to effectively leverage motion supervision in videos. The most critical ones are the number of segmentation channels $C$ and the object channel index $c_o$.

Previous work \cite{liu2021emergence} tunes $C$ with ground truth annotation on the validation set. \cite{liu2021emergence} tunes $c_o$ by calculating each channel's distribution of prediction locations across frames, which is only valid for tuning $c_o$ and has limited capability towards other hyperparams.

We found that while a good segmentation includes the whole object, an unsatisfying one either only includes parts of the foreground or includes the background. Therefore, we propose motion-semantic alignment as a metric, quantifying the segmentation quality. Similar to the definition in semantic constraint, we define semantic response $S_{t, c}$ as:

\vspace{-12pt}
\begin{equation}
S_{t, c} = \text{Threshold}(\sigma(f_\text{aux}(I_t), \text{GuidedPool}(f_\text{aux}(I_t), \hat M_{t, c})))
\end{equation}
where $\hat M_{t, c}$, the prediction from our model $g(f(\cdot))$ learned from motion supervision, is used as the query criterion, similar to \cref{sec:method_refinement}.
Then, we compute the IoU (Intersection-over-Union) between $\hat M_{t, c}$ and $S_{t, c}$ as the metric quantifying motion-semantic alignment, with an example in \cref{fig:tuning-example}. We then compare results with different hyperparam values and select the one with the highest mean IoU. % This method is applicable to general hyperparam in stage 1. Furthermore, we found that using only the first frame is enough for determining the key performance differences with different hyperparams, since the runs that do not fit well (e.g. split foreground into two parts or include background patches) can be easily detected. 

Our hyperparam tuning method is model-agnostic and applicable to other UVOS methods. We demonstrate its effectiveness by using it to tune $C$ and $c_o$, as in \cref{sec:experiment_tuning}.

\def\tabAblation#1{
\begin{table}[#1]
% \tablestyle{12.0pt}{1.1}
\setlength{\tabcolsep}{2pt}
\centering
\begin{tabular}{cccccl}
\shline
\begin{tabular}[c]{@{}c@{}}Residual\\pathway\end{tabular} &
\begin{tabular}[c]{@{}c@{}}Feature\\merging\end{tabular} & 
\begin{tabular}[c]{@{}c@{}}Low-level\\refinement\end{tabular} &
\begin{tabular}[c]{@{}c@{}}Semantic\\constraint\end{tabular}  & 
CRF & \multicolumn{1}{c}{$\mathcal{J}$ ($\uparrow$)} \\
\shline
       &        &        &        &        & 67.6 \\
\cmark &        &        &        &        & 73.0 {\small \textcolor{Green}{(+5.4)}} \\
       & \cmark &        &        &        & 71.0 {\small \textcolor{Green}{(+3.4)}} \\
\cmark & \cmark &        &        &        & 76.1 {\small \textcolor{Green}{(+8.5)}} \\
\cmark & \cmark & \cmark &        &        & 77.3 {\small \textcolor{Green}{(+9.7)}} \\
\cmark & \cmark & \cmark & \cmark &        & 77.7 {\small \textcolor{Green}{(+10.1)}} \\
\cmark & \cmark & \cmark & \cmark & \cmark & \textbf{79.8} {\small \textbf{\textcolor{Green}{(+12.2)}}} \\
\shline
\end{tabular}
\caption{\textbf{Effect of each component of our method (DAVIS16).} Residual pathway on its own provides the most improvement in our method. % Merging high-level and low-level feature allows sharper mask predictions. Both low-level refinement and semantic constraint allow us to use appearance supervision to correct the misconception from misleading motion signals. 
All components together contribute to $12.2\%$ gain.}
\label{tab:ablation}
\end{table}
}

\def\tabAblationResidual#1{
\begin{table}[#1]
% \tablestyle{12.0pt}{1.1}
\setlength{\tabcolsep}{17.5pt}
\centering
\vspace{-5pt}
\begin{tabular}{lc}
\shline
\multicolumn{1}{c}{Variants} & DAVIS16 $\mathcal{J}$ ($\uparrow$) \\
\shline
None & 67.6 \\
None (w/ robust loss \cite{sun2017pwc}) & 69.4 \\
Scaling & 70.4 \\
Residual (affine) & 68.8 \\
Residual & \textbf{73.0} \\
\shline
\end{tabular}
\vspace{-5pt}
\caption{\textbf{Ablations on additional pathway confirm our design choice of residual pathway.} We benchmark without feature merging or refinement stage to show the raw performance gain.\vspace{-5pt}}
\label{tab:ablation-residual}
\end{table}
}

\def\tabAblationCRF#1{
\begin{table}[#1]
% \tablestyle{12.0pt}{1.1}
\setlength{\tabcolsep}{10pt}
\centering
\begin{tabular}{lcc}
\shline
Ablation & Stage 1 & Stage 1 \& 2 \\
\shline
Without post-processing & 76.1 & 77.3 \\
With CRF post-processing & 78.4 & 79.5 \\
\hline
$\Delta$ & {\textcolor{Green}{+2.3}} & {\textcolor{Green}{+2.2}} \\
\shline
\end{tabular}
\caption{\textbf{The refinement CRF in our stage 2 is orthogonal to upsampling CRF in post-processing}, shown by negligible changes in performance improvements from post-processing w/ or w/o stage 2 (2.2 vs 2.3).\vspace{-5pt}}
\label{tab:ablation-crf}
\end{table}
}

\iffalse
\def\figTuning#1{
\begin{figure*}[#1]
\vspace{-10pt}
\begin{subfigure}[b]{0.5\linewidth}
  \centerline{\includegraphics[width=0.85\textwidth]{figures/plot_num_channels.pdf}}
\end{subfigure}\hfill
\begin{subfigure}[b]{0.5\linewidth}
  \centerline{\includegraphics[width=0.85\textwidth]{figures/plot_channel_index.pdf}}
\end{subfigure}
\vspace{-30pt}
\caption{\textbf{The tuned hyperparams from unsupervised motion-semantic alignment greatly resemble the ones obtained with human annotation.} We thus use $C=4$ segmentation channels for IMAS and object channel $c_o=3$ for this run. Although $c_o$ varies in each training run by design \cite{liu2021emergence}, our tuning method has negligible overhead and could be performed after training to find $c_o$ \textit{within seconds}.}
\label{fig:tuning}
\end{figure*}
}
\fi

\def\figTuning#1{
\begin{figure}[#1]
\vspace{-5pt}
\centerline{\includegraphics[width=1.0\linewidth]{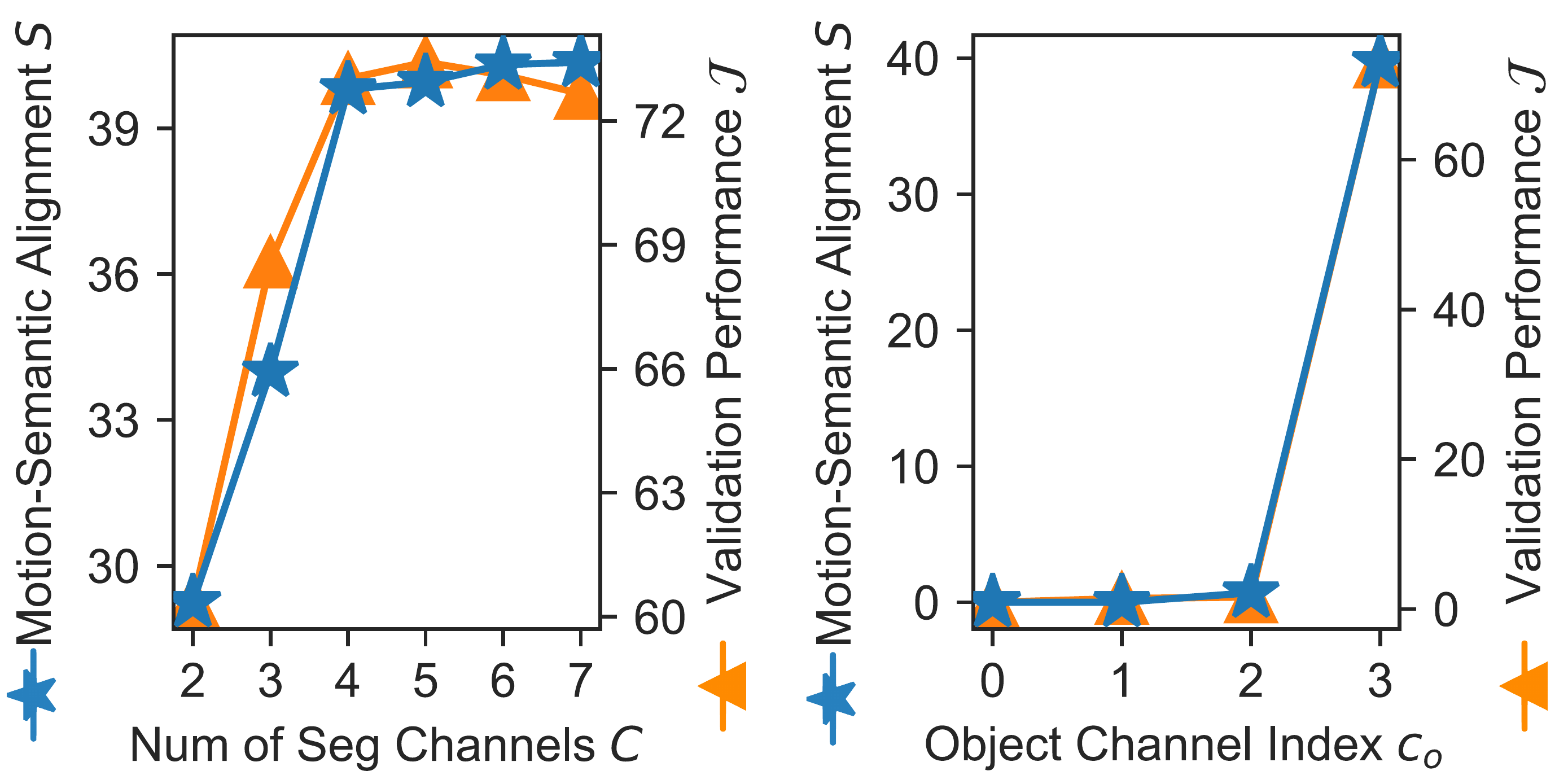}}
\vspace{-10pt}
\caption{\textbf{The tuned hyperparams from motion-semantic alignment align with the ones from human annotation.} We thus use $C=4$ segmentation channels for IMAS and object channel $c_o=3$ for this run. Although $c_o$ varies in each training run by design \cite{liu2021emergence}, our tuning method has negligible overhead and can be performed after training ends to find $c_o$ \textit{within seconds}.\vspace{-5pt}}
\label{fig:tuning}
\end{figure}
}

\def\figVisualizations#1{
\begin{figure*}[#1]
\vspace{-20pt}
\centering
\includegraphics[width=1.0\textwidth]{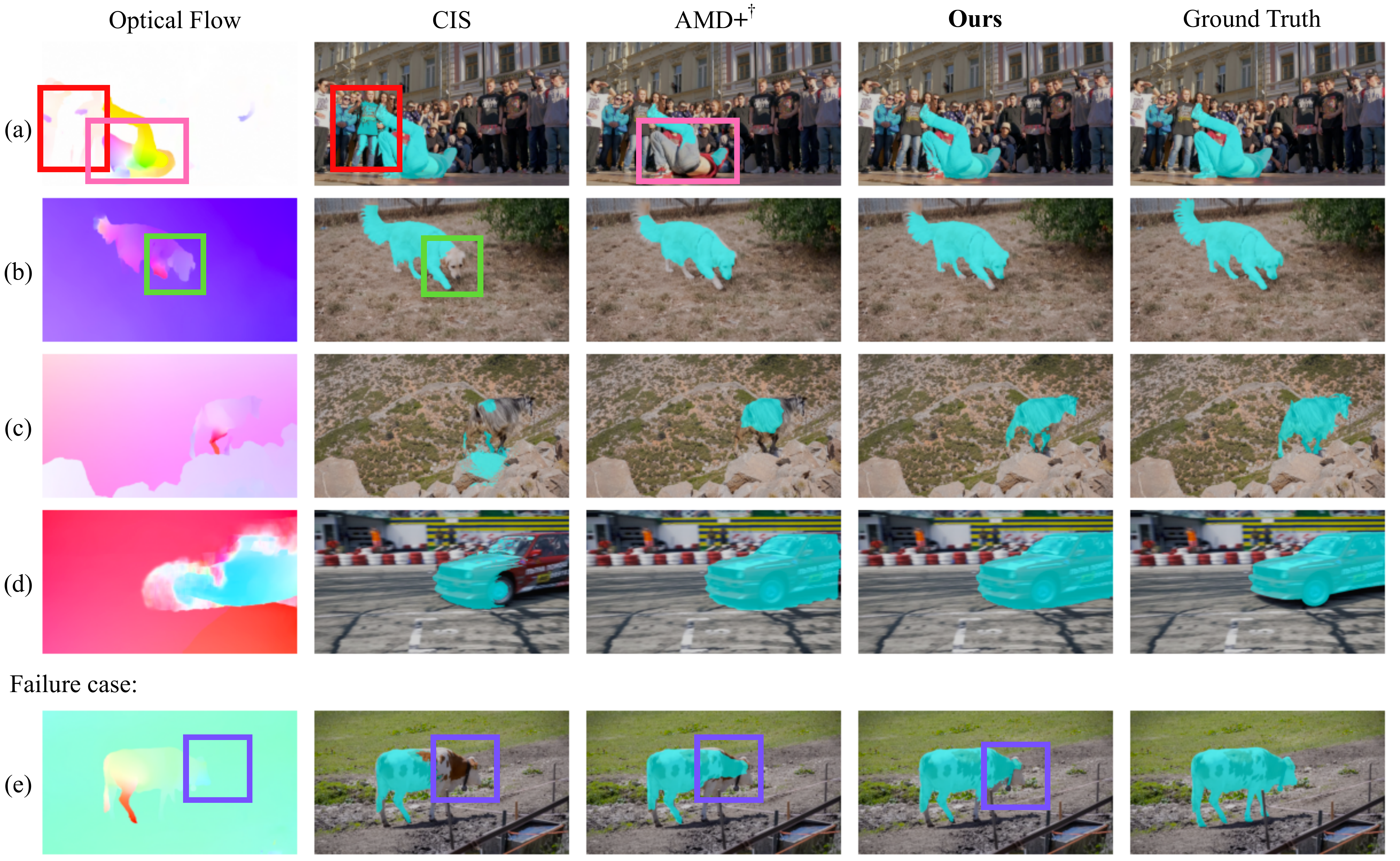}
\vspace{-15pt}
\caption{\textbf{Our method is robust to uninformative or even misleading motion cues.} Comparisons with CIS \cite{yang2019unsupervised} and AMD \cite{liu2021emergence} show our improvements in challenging scenes with complex foreground motion (a)(b), distracting background motion (a)(c), depth effect from camera motion (c), and fast camera/object motion (c)(d). The failure case (e) demonstrates a scenario where neither motion nor appearance information is informative, causing under-segmentation. $\dagger$ denotes AMD \cite{liu2021emergence} with higher-quality RAFT flow \cite{teed2020raft} for a closer comparison. \vspace{-15pt}}
\label{fig:visualizations}
\end{figure*}
}

\tabMainComparison{!t}
\section{Experiments}
\label{experiments}

\subsection{Datasets}
We evaluate our methods on three datasets commonly used to benchmark UVOS, following previous works \cite{yang2019unsupervised,yang2021self,liu2021emergence,choudhury2022guess,xie2022segmenting}. \textbf{DAVIS2016} \cite{perazzi2016benchmark} contains 50 video sequences of 3,455 frames. Performance is evaluated on validation set that includes 20 videos with annotation at 480p resolution. \textbf{SegTrackv2} (STv2) \cite{li2013video} contains 14 videos of different resolutions with 976 annotated frames with lower image quality than \cite{perazzi2016benchmark}. \textbf{FBMS59} \cite{ochs2013segmentation} contains 59 videos with 13,860 frames in total and 720 frames annotated with a roughly fixed interval. We follow previous work to merge multiple foreground objects in STv2 and FBMS59 into one mask and train on all unlabeled videos. We adopt mean Jaccard index~$\mathcal{J}$ (mIoU) as the main evaluation metric.

\subsection{Unsupervised Video Object Segmentation}
\noindent\textbf{Setup.} Our architecture is simple and straightforward. We use a ResNet50 \cite{he2016deep} backbone followed by a segmentation head and a residual prediction head. Both heads only consist of three Conv-BN-ReLU layers with 256 hidden units. This standard design allows efficient implementation for real-world applications. We use $C=4$ object channels by default, which is found without human annotation, as we will show in \cref{sec:experiment_tuning}. The object channel index $c_o$ is also determined in this manner. The RAFT \cite{teed2020raft} model that we use is only trained on synthetic FlyingChairs \cite{dosovitskiy2015flownet} and FlyingThings \cite{mayer2016large} dataset without human annotation. We refer readers to supp. materials for more details.

\noindent\textbf{Results.} As in \cref{tab:main-comparison}, IMAS outperforms previous methods under fair comparison, often by a large margin. On DAVIS16, IMAS surpasses the previous state-of-the-art method by $5.6\%$ without post-processing. With CRF as the only post-processing, IMAS improves on previous methods by $8.3\%$, despite competing methods employing techniques such as multi-step flow, multi-crop ensemble, and temporal smoothing. IMAS even surpasses the previous method that employs more complex Swin-T + MaskFormer architecture \cite{liu2021swin,cheng2021maskformer} by $5.4\%$ with post-processing. Despite the varying image quality in STv2 and FBMS59, IMAS improves over past methods under fair comparison, both without and with light post-processing. Semantic constraint could be included if additional gains are desired. However, IMAS still achieves competitive performance without semantic constraint, thus \textit{not relying on external frozen features}.

% Note: If possible, we can add this (in supplementary materials).
% \subsection{Zero-shot Video Object Segmentation}

% Note: Potentially compute correlation (in supplementary materials).
\subsection{Motion-Semantic Alignment for Hyperparam Tuning}
\label{sec:experiment_tuning}
We use motion-semantic alignment as a metric to tune two key hyperparameters, the number of segmentation masks $C$ and the object channel index $c_o$. As in \cref{fig:tuning}, despite \textit{not using any manual annotation}, we observe that using motion-semantic alignment is highly effective for hyperparam tuning due to its high correlation to downstream performance evaluated with ground truth. For \textbf{the number of segmentation masks $C$}, we found that increasing the number of channels improves the segmentation quality of our model by increasing the fitting power. However, such an increase saturates at $C=4$. Therefore, we use $C=4$ for all experiments unless otherwise stated.

For \textbf{the object channel index $c_o$}, since $c_o$ changes w.r.t random initialization by design\cite{liu2021emergence}, optimal $c_o$ needs to be obtained at the end of each training run. We propose to leverage the redundancy in video sequences and use only the first frame of each video sequence for finding $c_o$. With this adjustment, our tuning method could complete within \textit{only 3 seconds} for each candidate channel, which allows our tuning method to be performed after the whole training run with negligible overhead to find the object channel $c_o$. 

\figTuning{!t}
\figVisualizations{!t}

\subsection{Ablation Study}
\label{sec:ablation}

\tabAblation{!t}
\tabAblationResidual{!t}
\tabAblationCRF{!t}

\noindent\textbf{Contributions of each component.} As in \cref{tab:ablation}, residual pathway allows more flexibility and contributes $5.4\%$. Feature merging indicates concatenating the feature from the first block and the last block of ResNet for higher feature resolution ($96\!\times\!96$/$98\!\times\!175$ in training/inference), which allows an additional $3.4\%$ improvement with negligible increase in model size. Note that our prediction resolution after feature merging is still lower than most previous works (e.g., $128\!\times\!224$ in \cite{yang2021self} and $192\!\times\!384$ in \cite{yang2019unsupervised}) and thus our performance gain to other methods is \textit{not} due to higher output resolutions. The explicit appearance refinement from the second stage boosts the performance to $77.7\%$, with a $10.1\%$ gain in total. The CRF post-processing leads to $79.8\%$ in mean Jaccard index, a $12.2\%$ increase to baseline.

\noindent\textbf{Designing additional pathway.} In \cref{tab:ablation-residual}, we show that robustness loss \cite{sun2017pwc,liu2020learning} does not effectively relieve from misleading motion. We also implemented a pixel-wise scaling pathway, which multiplies each value of the motion vector by a predicted value. Furthermore, we fit an affine transformation per segmentation channel as the residual. The pixel-wise residual performs the best in our setting and is chosen in our model, showing the effectiveness of a \textit{learnable} and \textit{flexible} motion model inspired by relative motion.

\noindent\textbf{Orthogonality of our appearance supervision with post-processing}. The refined masks after our appearance-based refinement have the same resolution as the original exported masks. Therefore, the refinement CRF in stage 2 has an orthogonal effect to the upsampling CRF in post-processing that is mainly used to create high-resolution masks from bilinearly-upsampled ones. As shown in \cref{tab:ablation-crf}, the gains that come from post-processing do not diminish as we apply appearance-based refinement in stage 2, which also shows the orthogonality of our refinement to post-processing.

\subsection{Visualizations and Discussions}
In \cref{fig:visualizations}, we compare IMAS with \cite{yang2019unsupervised,liu2021emergence}. Our method adapts to challenging cases such as complex non-uniform foreground motion, distracting background motion, and camera motion including rotation. However, IMAS still has limitations: IMAS does not work if neither motion nor appearance provides informative signals and thus is misled by the texture of the cow in \cref{fig:visualizations}(e). Although IMAS has the ability to recognize multiple foreground objects when moved in sync, it sometimes learns to capture only one object when the objects move in very different patterns. Finally, IMAS is not designed to separate multiple foreground objects. More visualizations/discussions are in supp.~mat.

\section{Summary}
\label{summary}
We present IMAS, an unsupervised video object segmentation method by leveraging motion-appearance synergy. Our method has an object discovery stage with a conflict-resolving learnable residual pathway and a refinement stage with appearance supervision. We also propose motion-semantic alignment as an annotation-free hyperparam tuning method. Extensive experiments show our effectiveness and utility in challenging cases. % in future research and applications.

\def\figVisualizationsSuppDAVIS#1{
\begin{figure*}[#1]
\centering
\includegraphics[width=1.0\textwidth]{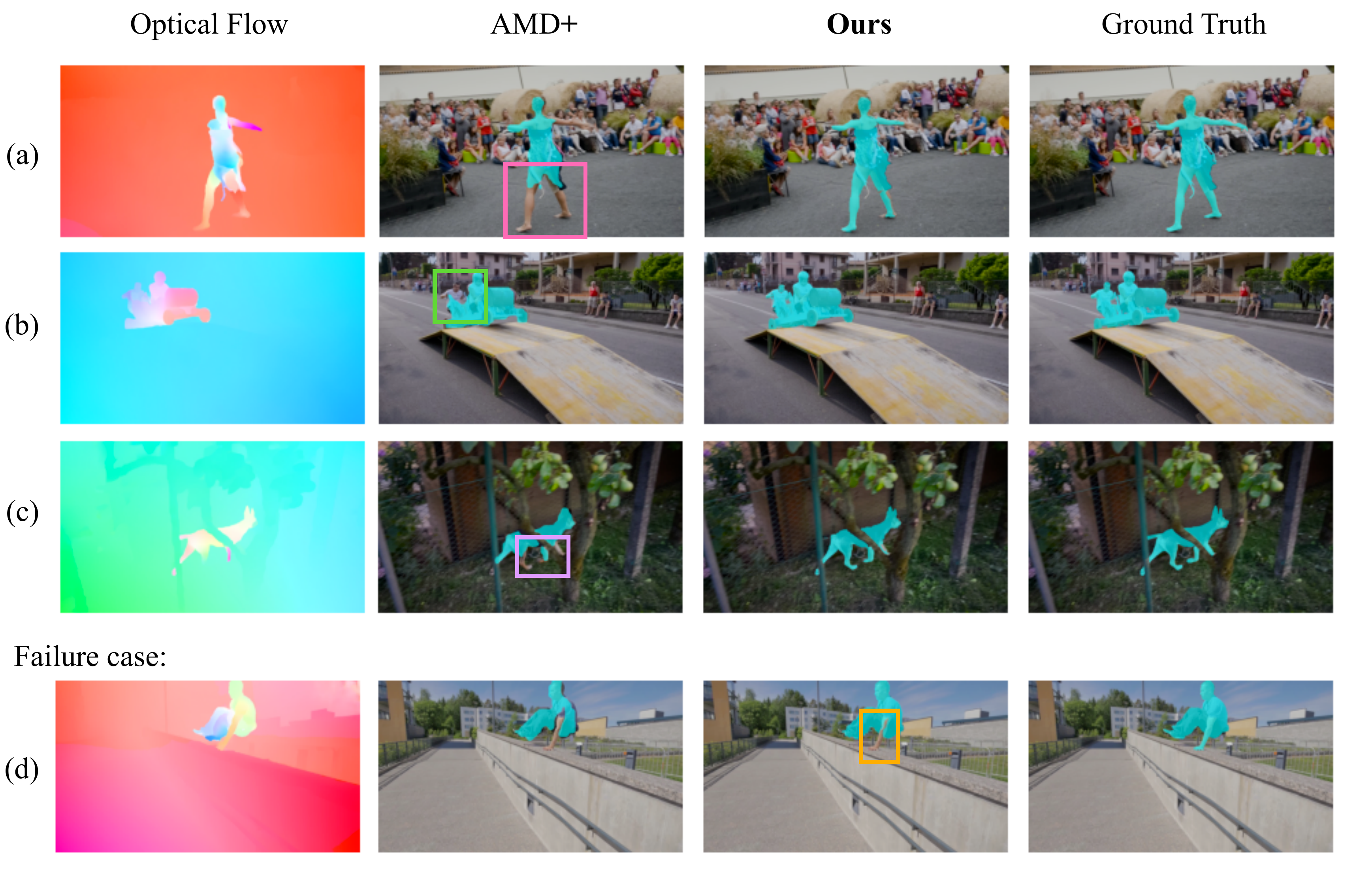}
\caption{\textbf{Additional visualizations on DAVIS16 \cite{perazzi2016benchmark}.} Our method is much more robust to scenes when motion and appearance conflict (a-c). Our method also segments out multiple foreground objects as foreground when they move together, as consistent to what humans perceive (b). However, when neither motion or appearance is informative, our method may be mislead to exclude part of an object (d).}
\label{fig:visualizations_davis}
\end{figure*}
}

\def\figVisualizationsSuppST#1{
\begin{figure*}[#1]
\centering
\includegraphics[width=1.0\textwidth]{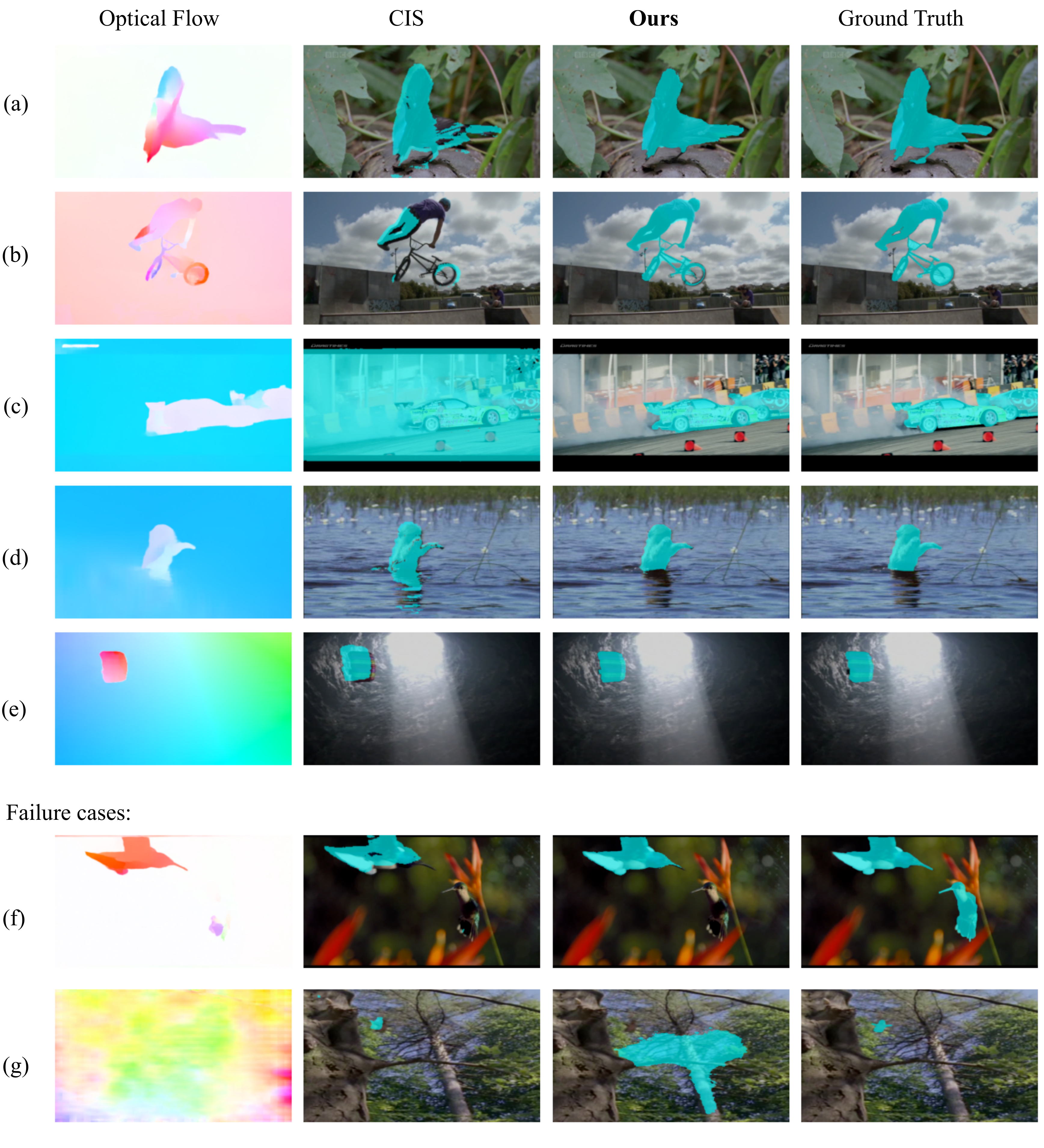}
\caption{\textbf{Additional visualizations on STv2 \cite{li2013video}.} Our method, with the residual flow, could model the non-uniform 2D flow from object rotation in 3D (a), as long as the rotation flow is within our constraint upper bound in residual flow. Our method also captures multiple objects in a foreground group (b)(c). Our method is robust to misleading common motion naturally occurred in real-world scenes (d). Finally, our method is robust to the camera motion that leads to non-uniform background flow (e). However, our method may select to focus on only one of the foreground objects if one has significantly larger motion than the other (f). Our method still could not work in scenes with misleading motion and complicated appearance unable to be parsed by the model (g).}
\label{fig:visualizations_st}
\end{figure*}
}

\def\figVisualizationsSuppFBMS#1{
\begin{figure*}[#1]
\centering
\includegraphics[width=1.0\textwidth]{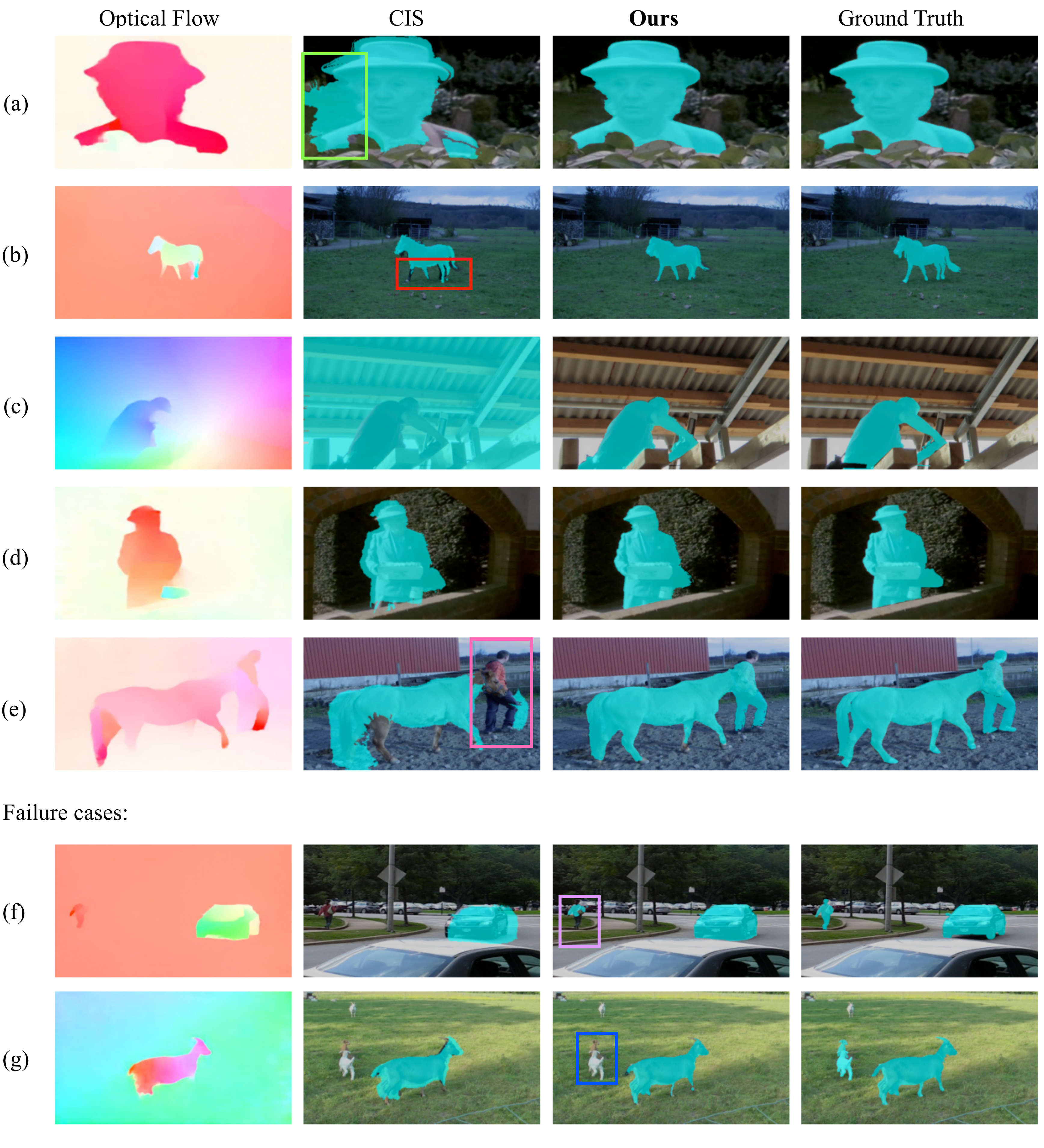}
\caption{\textbf{Additional visualizations on FBMS59 \cite{brox2010freiburg,ochs2013segmentation}.} Our method is robust to scenes with complicated appearance distractions (a). Our method also works with fine details (b) as well as camera motion and adjustment (\eg, zooming in) (c). Our method also segments out multiple foreground objects with coherent movement (e). However, when multiple objects exist and one moves significantly faster than the other, our method sometimes focuses on the object with larger movement, setting them as the only foreground (f) (g). Note that the ground truth also comes with ambiguity in (g), as the goat in the back is not annotated while the goats in the front are in the annotation.}
\label{fig:visualizations_fbms59}
\end{figure*}
}

\def\figVisualizationsSuppResidual#1{
\begin{figure*}[#1]
\centering
\includegraphics[width=1.0\textwidth]{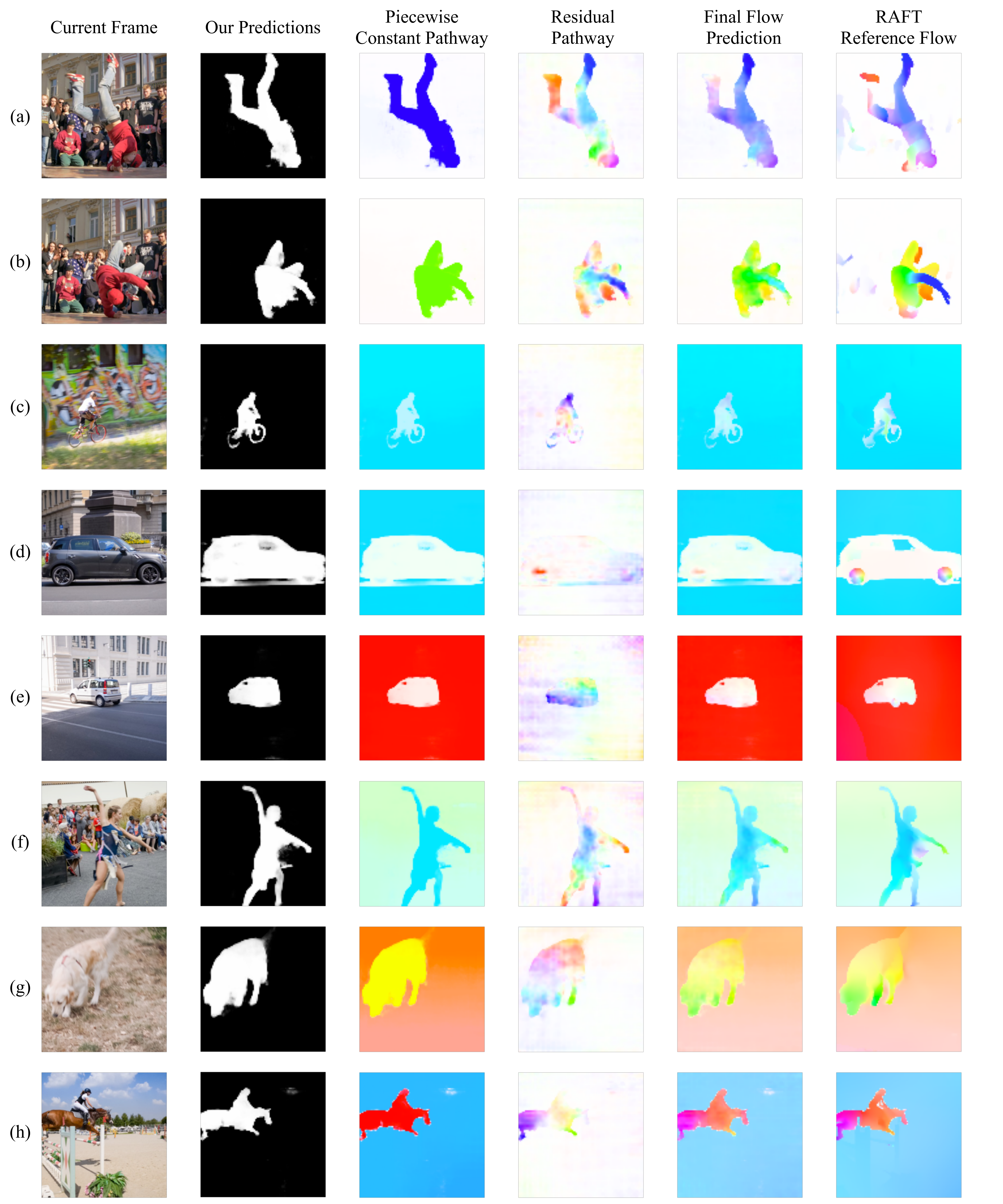}
\caption{\textbf{Visualizations for both pathways} show that the introduction of residual pathway allows our segmentation prediction to better align with the appearance cues rather than exactly correspond to the motion signals, leading to high quality segmentation especially when appearance and motion cues conflict. Modeling relative motion and other non-uniform motion patterns in 2D flow such as the depth effect, residual pathway makes our method flexible and robust to objects with complex motion by leveraging synergy in motion and appearance.}
\label{fig:visualizations_residual}
\end{figure*}
}

\def\figTuningAMD#1{
\begin{figure}[#1]
\centerline{\includegraphics[width=1.0\linewidth]{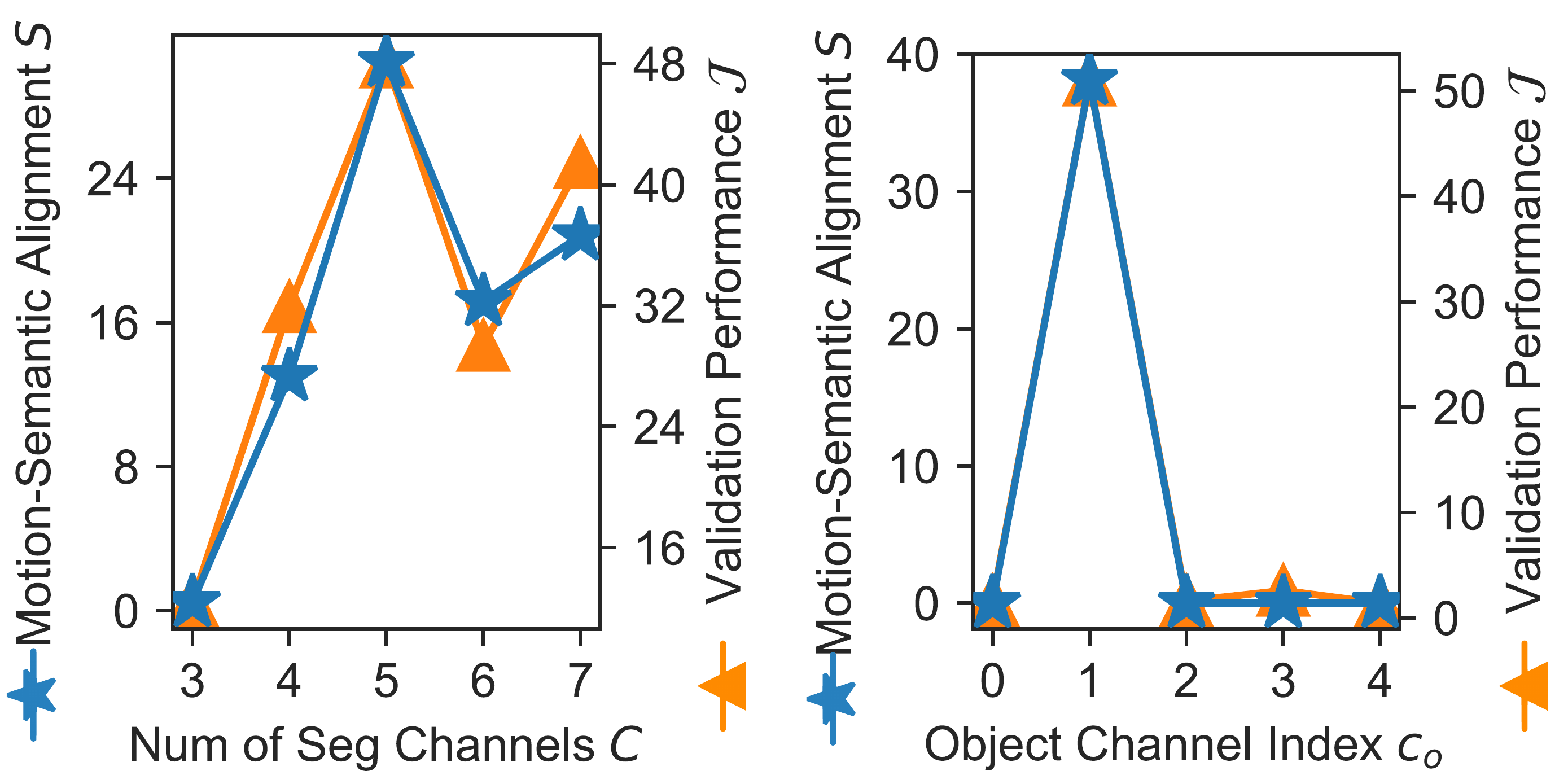}}
\caption{\textbf{Our hyperparam tuning technique is model agnostic.} When using our model-agnostic hyperparam tuning technique on AMD \cite{liu2021emergence}, the tuned hyperparams from unsupervised motion-semantic alignment greatly resemble the ones obtained with human annotation. In this training run, the number of segmentation channels is 4 to be optimal, and the object channel index is 1 from \textit{both} our motion-semantic alignment and validation performance. Although $c_o$ varies in each training run by design \cite{liu2021emergence}, our tuning method has negligible overhead and could be performed after training to find $c_o$ \textit{within seconds}.
}
\label{fig:tuning_amd}
\end{figure}
}

\def\tabDAVISPerSequence#1{
\begin{table}[#1]
% \tablestyle{12.0pt}{1.1}
\setlength{\tabcolsep}{20pt}
\centering
\begin{tabular}{lc}
\shline
Sequence & $\mathcal{J}$ \\
\hline
blackswan & 66.1 \\
bmx-trees & 60.2 \\
breakdance & 81.0 \\
camel & 83.9 \\
car-roundabout & 87.6 \\
car-shadow & 85.3 \\
cows & 87.0 \\
dance-twirl & 85.7 \\
dog & 82.8 \\
drift-chicane & 78.8 \\
drift-straight & 78.8 \\
goat & 80.1 \\
horsejump-high & 87.4 \\
kite-surf & 64.2 \\
libby & 82.8 \\
motocross-jump & 72.0 \\
paragliding-launch & 63.0 \\
parkour & 86.2 \\
scooter-black & 82.1 \\
soapbox & 87.2 \\
\hline
Frame Avg & 79.8 \\
\shline
\end{tabular}
\caption{\textbf{Per sequence Jaccard index $\mathcal{J}$ on DAVIS16 \cite{perazzi2016benchmark}.}}
\label{tab:davis16_per_seq}
\end{table}
}

\def\tabSTPerSequence#1{
\begin{table}[#1]
% \tablestyle{12.0pt}{1.1}
\setlength{\tabcolsep}{20pt}
\centering
\begin{tabular}{lc}
\shline
Sequence & $\mathcal{J}$ \\
\hline
bird of paradise & 86.7 \\
birdfall & 1.9 \\
bmx & 75.1 \\
cheetah & 50.6 \\
drift & 87.1 \\
frog & 77.1 \\
girl & 83.2 \\
hummingbird & 54.6 \\
monkey & 86.4 \\
monkeydog & 15.4 \\
parachute & 93.8 \\
penguin & 64.5 \\
soldier & 82.0 \\
worm & 85.6 \\
\hline
Frame Avg & 72.7 \\
\shline
\end{tabular}
\caption{\textbf{Per sequence Jaccard index $\mathcal{J}$ on STv2 \cite{li2013video}.}}
\label{tab:stv2_per_seq}
\end{table}
}

\def\tabFBMSPerSequence#1{
\begin{table}[#1]
% \tablestyle{12.0pt}{1.1}
\setlength{\tabcolsep}{20pt}
\centering
\begin{tabular}{lc}
\shline
Sequence & $\mathcal{J}$ \\
\hline
camel01 & 89.8 \\
cars1 & 85.4 \\
cars10 & 32.4 \\
cars4 & 83.9 \\
cars5 & 80.0 \\
cats01 & 90.4 \\
cats03 & 69.8 \\
cats06 & 56.9 \\
dogs01 & 61.7 \\
dogs02 & 84.4 \\
farm01 & 80.5 \\
giraffes01 & 58.5 \\
goats01 & 78.1 \\
horses02 & 82.4 \\
horses04 & 77.3 \\
horses05 & 53.5 \\
lion01 & 78.1 \\
marple12 & 76.3 \\
marple2 & 85.4 \\
marple4 & 87.2 \\
marple6 & 54.4 \\
marple7 & 78.6 \\
marple9 & 49.6 \\
people03 & 81.2 \\
people1 & 79.8 \\
people2 & 83.6 \\
rabbits02 & 74.0 \\
rabbits03 & 55.8 \\
rabbits04 & 0.5 \\
tennis & 78.5 \\
\hline
Frame Avg & 69.0 \\
\shline
\end{tabular}
\caption{\textbf{Per sequence Jaccard index $\mathcal{J}$ on FBMS59 \cite{brox2010freiburg,ochs2013segmentation}.}}
\label{tab:fbms59_per_seq}
\end{table}
}

\FloatBarrier
\figVisualizationsSuppResidual{!htbp}
\figVisualizationsSuppDAVIS{!htbp}
\figVisualizationsSuppST{!htbp}
\figVisualizationsSuppFBMS{!htbp}

\section{Appendix}
\label{appendix}
\subsection{Additional Visualizations and Discussions}
We present additional visualizations on the three main datasets that we benchmark on \cite{perazzi2016benchmark,li2013video,brox2010freiburg,ochs2013segmentation}. We demonstrate high-quality segmentation in several challenging cases. We also discuss some limitations of our method with examples.

\subsubsection{Visualizations of Residual Pathway}
As in \cref{fig:visualizations_residual}, the introduction of residual pathway allows our segmentation prediction to better align with the appearance cues rather than exactly correspond to the motion signals, which leads to high quality segmentation especially when appearance and motion cues conflict. Modeling relative motion and other non-uniform motion patterns in 2D flow such as the depth effect, residual pathway makes our method flexible and robust to objects with complex motion by leveraging synergy in motion and appearance.

\subsubsection{DAVIS2016}
As in \cref{fig:visualizations_davis}, our method is much more robust to scenes when motion and appearance conflict and could segment out multiple foreground objects as foreground when they move together, as consistent with human perception. However, when neither motion nor appearance is informative, our method may be mislead to exclude part of an object.

\subsubsection{SegTrackv2}
As in \cref{fig:visualizations_st}, our method could model the non-uniform 2D flow from object rotation in 3D with residual flow. Our method also captures multiple objects in a foreground group when they share similar motion. Our method is robust to misleading common motion naturally occurred in real-world scenes such as reflections and to the camera motion that leads to non-uniform background flow. However, our method may select to focus on only one of the foreground objects if one has significantly larger motion than the other. Our method still could not work in scenes with misleading motion and complicated appearance unable to be parsed by the model.

\subsubsection{FBMS59}
As in \cref{fig:visualizations_fbms59}, our method is robust to scenes with complicated appearance distractions. Our method also works with fine details as well as camera motion and adjustment (\eg, zooming in). Our method also segments out multiple foreground objects with coherent movement. However, when multiple objects exist and one moves significantly faster than the other, our method sometimes focuses on the object with larger movement, setting them as the only foreground.

% Potential: with residuals and flow.
% Small objects
% Non-uniform movement (e.g. rotation)
% Parts

\subsection{Applying Motion-Semantic Alignment on Previous Work}
\figTuningAMD{htbp}

To show that our hyperparam tuning method with motion-semantic alignment is \textit{model-agnostic}, we also apply our hyperparam tuning method to AMD \cite{liu2021emergence} to find the number of segmentation channels $C$ and the channel index $c_o$. We follow the original setting in AMD and train on Youtube-VOS \cite{xu2018youtube} without test adaptation. % As in the semantics matching process for our model, we do not consider samples with large background overlap (\ie, having more than 80\% of the image width or 90\% of the image height) by zeroing the metric value for such match instances.

As in \cref{fig:tuning_amd}, our tuning method also works on AMD to find the number of segmentation channels and the object channel index, showing that our tuning method is model-agnostic. Our method gives a metric that significantly correlates with the validation performance, despite having a different scale, without human annotation. Our method finds the same optimal number of channels $C$ and object channel index $c_o$ as using the validation set performance with human annotation.

% To show that our hyperparam tuning method with motion-semantic alignment is model-agnostic, we also apply our hyperparam tuning method to AMD \cite{liu2021emergence} to tune the number of channels $C$ and to find the channel index $c_o$. Note that the optimal $C$ may vary between methods and $c_o$ changes with initialization.

% AMD doesn't have residual so it is reasonable to expect the optimal number of channels to be higher due to lower tolerance on deviation from piecewise constant motion per channel.

Compared to constraining the object to a certain channel with regularization on the loss so that $c_o$ is fixed, as used in \cite{choudhury2022guess}, our method with motion-semantic alignment does not require changing the implementation on training and could be performed offline with a trained model. Our method does not introduce additional hyperparams to balance the regularization factor either.

\subsection{Additional Implementation Details}
Our setting mostly follows previous works \cite{liu2021emergence,choudhury2022guess}. Following the implementation in \cite{liu2021emergence}, we treat video frame pair $\{t, t+1\}$ as both a forward action from time $t$ to time $t+1$ and a backward action from time $t+1$ and $t$, since they follow similar rules. Therefore, we use this to implement a symmetric loss that applies the loss function both forward and backward and sum them up to get the final loss. Note that this could be understood as a data augmentation that always supplies a pair in forward and backward to the training batch. However, since our ResNet shares weights for each image input, the feature for each input is reused by forward and backward action, thus the symmetric loss only adds marginal computation and is included in our implementation as well. 

Furthermore, following \cite{liu2021emergence}, for DAVIS16, we use random crop augmentation at training to crop a square image from the original image. At test time, we directly input the original image (which is non-square). Note that the augmentation makes the image size different for training and test, but as ResNet\cite{he2016deep} takes images of different sizes, this does not pose a problem empirically. In STv2 and FBMS59, the images have very different aspect ratios (some having a height lower than the width) and thus a square crop is no longer feasible to obtain the same size of images for a batch unless we resize. Therefore, we perform resize rather than random crop to make all images the size of 480p before the standard pipeline. However, this reduces the randomness in augmentation and may lead to loss in performance, which remains to be addressed as an implementation decision for the future to further improve the performance. We additionally use pixel-wise photo-metric transformation \cite{mmseg2020} for augmentation with default hyperparam. 

As for the architecture, we found that simply adding a convolutional head to the last layer feature in ResNet provides insufficient detailed information for high-quality output. Rather than incorporate a heavy segmentation head (\eg, \cite{cheng2021maskformer} in \cite{choudhury2022guess}), to keep our architecture easy to implement, we only change the head in a simple fashion by fusing feature from the first residual block and the third residual block in ResNet. This allows the feature to jointly capture high-level information and low-level details, which is termed feature merging. Since the backbone remains the same, feature merging adds negligible compute. Note that feature merging is only in segmentation head, and residual prediction is simply bilinearly upsampled. Due to lower image resolution, no feature merging is performed for STv2. Following \cite{choudhury2022guess}, we load self-supervised ImageNet pretrained weights \cite{ILSVRC15,wang2021dense} learned without annotation, as the training video datasets are too small for learning generalizable feature (\eg, DAVIS16/STv2/FBMS59 has only 3,455/976/13,860 frames). We observe that image-based training can be replaced by training on uncurated Youtube-VOS \cite{xu2018youtube} with our training process, as in \cite{liu2021emergence}, so that one implementation is used throughout training for simplicity in the real-world applications.

For matching the semantics with frozen feature \cite{wang2021dense} in creating the semantic constraint, to prevent using the background to filter out the foreground, we discard the potential constraint if the match is greater than 80\% of the width or 90\% of the height, since it is likely that we match the background given such condition. Such a rule also applies to the semantic matching for hyperparam tuning, in which we set IoU for the current frame to be zero if the match follows the same condition. The dilation to process the mask in semantic constraint is 3x3 on DAVIS16 and 7x7 on STv2 and FBMS59 due to lower image quality.

In our training, we follow \cite{liu2021emergence} and use batch size 16 (with two images in a pair, and thus 32 images processed in each forward pass). Stage 1 and stage 2 take around 40k and 200 iterations, respectively, for DAVIS16. We use learning rate $1\times10^{-4}$ with Adam optimizer \cite{kingma2014adam} and polynomial decay (factor $0.9$, min learning rate $1\times10^{-6}$). We set weight decay to $1\times10^{-6}$. In stage 2, we set $w_\text{motion}=0.1$ and $w_\text{app}=2.0$ to balance the two losses.

\subsection{Per-sequence Results}
\tabDAVISPerSequence{p}
\tabSTPerSequence{p}
\tabFBMSPerSequence{p}

We list our per-sequence results on DAVIS16 \cite{perazzi2016benchmark}, STv2\cite{li2013video}, FBMS59\cite{brox2010freiburg,ochs2013segmentation} in \cref{tab:davis16_per_seq}, \cref{tab:stv2_per_seq}, and \cref{tab:fbms59_per_seq}, respectively.

\subsection{Future Directions}
As our method does not impose temporal consistency, information redundancy from neighboring frames is not leveraged effectively. Our method could be more robust by using such information to deal with frames with insufficient motion or appearance information. Temporal consistency measures such as matching warped predictions as a regularization could be taken care of with an additional loss term or with post-processing as in \cite{yang2019unsupervised}.

Furthermore, our method currently does not support segmenting multiple parts of the foreground. Methods such as normalized cuts \cite{shi2000normalized} could be used to split the foreground into a number of objects with motion and appearance input for training the model with bootstrapping to provide training or correction signals.

\FloatBarrier

\clearpage

%%%%%%%%% REFERENCES
{\small
\bibliographystyle{ieee_fullname}
\bibliography{egbib}
}

\end{document}